\documentclass[nohyperref]{article}

\usepackage{hyperref}
\usepackage{url}

\usepackage{graphicx,wrapfig}
\usepackage{algpseudocode}
\usepackage{microtype}
\usepackage{graphicx}
\usepackage{subfigure}
\usepackage{float}

\usepackage{booktabs} % for professional tables
\usepackage{mathtools}
\usepackage{hyperref}
\usepackage{amsmath}
\usepackage{amssymb}
\usepackage{mathtools}
\usepackage{amsthm}
\usepackage[capitalize,noabbrev]{cleveref}
\theoremstyle{plain}
\newtheorem{theorem}{Theorem}[section]

\theoremstyle{definition}
\newtheorem{definition}[theorem]{Claim}

\theoremstyle{remark}

\usepackage[accepted]{icml2022}

\icmltitlerunning{Representative Subset Selection for Efficient Fine-Tuning in Self-Supervised Speech Recognition}

\begin{document}

\twocolumn[
\icmltitle{Representative Subset Selection for Efficient Fine-Tuning \\ in Self-Supervised Speech Recognition}




\begin{icmlauthorlist}
\icmlauthor{Abdul Hameed Azeemi}{xxx}
\icmlauthor{Ihsan Ayyub Qazi}{xxx}
\icmlauthor{Agha Ali Raza}{xxx}
\end{icmlauthorlist}

\icmlaffiliation{xxx}{Lahore University of Management Sciences}

\icmlcorrespondingauthor{Abdul Hameed Azeemi}{21030027@lums.edu.pk}

\icmlkeywords{Machine Learning, ICML}

\vskip 0.3in
]



\printAffiliationsAndNotice{} % otherwise use the standard text.

\begin{abstract}

% Abstracts must be a single paragraph, ideally between 4--6 sentences long.
% Gross violations will trigger corrections at the camera-ready phase.

Self-supervised speech recognition models require considerable labeled training data for learning high-fidelity representations for Automatic Speech Recognition (ASR) which is computationally demanding and time-consuming. We consider the task of identifying an optimal subset of data for efficient fine-tuning in self-supervised speech models for ASR. We discover that the dataset pruning strategies used in vision tasks for sampling the most informative examples do not perform better than random subset selection on fine-tuning self-supervised ASR. We then present the \textsc{Cowerage} algorithm for representative subset selection in self-supervised ASR. \textsc{Cowerage} is based on our finding that ensuring the coverage of examples based on training Word Error Rate (WER) in the early training epochs leads to better generalization performance. Extensive experiments with the wav2vec 2.0 and HuBERT model on TIMIT, Librispeech, and LJSpeech datasets show the effectiveness of \textsc{Cowerage} and its transferability across models, with up to 17\% relative WER improvement over existing dataset pruning methods and random sampling. We also demonstrate that the coverage of training instances in terms of WER values ensures the inclusion of phonemically diverse examples, leading to better test accuracy in self-supervised speech recognition models.

\end{abstract}

\section{Introduction}
\label{submission}

There has been rapid progress in recent years toward improving speech self-supervised learning (speech SSL) models. Such models learn high-fidelity speech representations using a large amount of unlabeled data and use paired data for fine-tuning on the downstream task of automatic speech recognition (ASR) \citep{baevski2020wav2vec, hsu2021hubert}. However, still a significant amount of labeled training data is used in the fine-tuning step, which is computationally demanding and time-consuming. For example, the standard \texttt{wav2vec2} fine-tuning procedure on Librispeech/Libri-light requires $\sim$$50$$-$$100$\,hours on a V100 GPU, which is significantly higher ($> 50\times$) than the cost of fine-tuning BERT on GLUE  \cite{lai2021parp}. Moreover, this also hinders their usage in low-resource systems, especially compute-restricted environments (e.g., cheaper GPUs and on-device computing), which is presently a significant barrier in democratizing access to these models \citep{ahmed2020democratization, paul2021deep}. 

Recent work uses adapters to enable efficient fine-tuning by using a fraction of parameters in speech SSL models \cite{thomas2022efficient}. However, their usage necessitates task-specific modifications, which prevents their applicability across different models and datasets. In contrast, we consider increasing the efficiency of speech SSL fine-tuning procedure by reducing training data requirements and find \textit{smaller}, \textit{representative} and \textit{model-agnostic} subsets of data for fine-tuning speech SSL models. Finally, we consider how example diversity within optimal subsets affects generalization in speech SSL, which is an important theoretical question requiring further investigation.

The data pruning mechanisms specifically tailored for deep learning models have been studied extensively for standard vision tasks. These methods focus on selecting the most informative training examples \citep{toneva2018empirical, coleman2019selection, paul2021deep, raju2021accelerating, karamcheti2021mind, margatina-etal-2021-active, mindermann2022prioritized} which has been shown to perform better than the random selection of the training data. The methods for identifying the important examples in these cases are based on scores that are directly derived from the training properties and example difficulty such as the error vector norm \citep{paul2021deep}, the number of times an example is forgotten during training \citep{toneva2018empirical} or the holdout loss \cite{mindermann2022prioritized}. However, no such mechanism has been studied yet for data pruning in speech SSL models.

Studying the impact of the data subset selection on ASR model performance raises several questions: Can we identify a model-agnostic scoring method based on the training properties for dataset pruning in speech SSL without significantly sacrificing the test accuracy? What are the phoneme distributions of \textit{good} subsets of training data, and how do they affect the latent representations within speech SSL models? Can we analyze the training landscape of speech SSL and extract novel insights that can benefit other speech tasks? The answers to these questions will help construct representative subsets that will benefit the paradigm of optimal dataset construction.

We find that in standard datasets for training speech SSL models, sampling only the \textit{hard-to-learn} training examples based on word error rate (WER)  does not consistently perform better than random pruning. This is in contrast to data pruning strategies in vision tasks where this method outperforms other baselines \citep{toneva2018empirical, paul2021deep, sorscher2022beyond}. For better data subset selection in fine-tuning speech SSL models, we propose \textsc{Cowerage}, an algorithm  designed to identify training examples important for better generalization. We find that ensuring the coverage of diverse examples based on \textit{training WER values} in the early training epochs leads to better accuracy on unseen test data than random pruning or selecting only the most informative (hard-to-learn) examples. Experiments show the effectiveness of the \textsc{Cowerage} algorithm over three primary pruning strategies: random selection, top \textit{k} (hardest subset selection), and bottom \textit{k} (easiest subset selection). 
To understand the underlying mechanism governing \textsc{Cowerage}'s generalization properties, we establish a connection between the training WER of the examples and their phonemic cover and find that our algorithm ensures the inclusion of phonemically diverse examples (i.e., examples of both low and high phonemic coverage) \textit{without} explicitly learning any phoneme-level error model. Finally, we demonstrate that phonemic diversity affects discrete latent representation within speech SSL, leading to performance gains via \textsc{Cowerage} subset selection.

\subsection{Our Contributions}
\begin{itemize}
    \item We use the WER of the \textit{individual training examples} as the basis for subset selection algorithms that prune the training data for speech SSL models (Section \ref{section:method}).

    \item We present \textsc{Cowerage}, an algorithm for selecting a subset of ASR training data that ensures uniform coverage of training WER values via a stratified random sampling approach (Section \ref{section:strategy3}). 

    \item Empirical evaluation on two models ---
 \texttt{wav2vec2} \citep{baevski2020wav2vec} and \texttt{HuBERT} \citep{hsu2021hubert} --- across three speech datasets --- TIMIT \citep{garofolo1993darpa}, Librispeech \citep{panayotov2015librispeech} and LJSpeech \citep{ito2017lj} --- show that fine-tuning on the subset selected by \textsc{Cowerage} gives a lower WER on the test split as compared to three other pruning strategies: random, top \textit{k}, and bottom \textit{k} examples (Section \ref{section:empiricaleval}). Additionally, we demonstrate that the subsets constructed through one model can be used for fine-tuning another speech SSL model, i.e., they are \textit{transferable} (Section \ref{section:transferabilityofsubsets}).

    \item We study the properties of the subsets selected by \textsc{Cowerage} by examining the phonemic coverage of training examples. We find that by ensuring the coverage of training WER values, \textsc{Cowerage} is able to select phonemically diverse examples, which results in a richer training subset (Section \ref{section:connectiontophonemes}). Finally, we establish the relationship between phonemic diversity and the discrete latent representation within speech SSL which allows \textsc{Cowerage} to perform better than random subset selection and hardest/easiest example selection (Section \ref{section:latentrepresentation}).

\end{itemize}

\section{Preliminaries}

Consider a self-supervised model $f(x ; \theta)$ ($\theta \in \mathcal{R}^{d}$) that is pre-trained on a large unlabelled dataset $x \in \mathcal{D}_{u}$ on some objective $\mathcal{L}_{p}$. The model obtained after self-supervised pretraining with weights $\theta_{L}$ is then fine-tuned for the downstream task of ASR with another objective $\mathcal{L}_{f}$ on a labelled dataset $x \in \mathcal{D}_{l}$ (which is generally smaller than $\mathcal{D}_{u}$). $\mathcal{D}_{l}$ consists of transcribed audios (i.e. audio and the corresponding sentence that was uttered). Our goal is to prune $\mathcal{D}_{l}$ to obtain a subset $ {B}_{l}$ such that the performance of self-supervised ASR model $f(x ; \theta)$ after fine-tuning on $ {B}_{l}$ is better than random pruning. We only consider pruning  $\mathcal{D}_{l}$ (and not $ {D}_{u}$) since we aim to directly evaluate the impact of different subset selection methods on the downstream task of ASR instead of the unsupervised pre-training of speech SSL model. The performance of an ASR model is commonly evaluated via WER ($\frac{I+D+S}{N}$), which is computed by aligning the word sequence generated by the ASR system with the actual transcription (containing \textit{N} words) and calculating the sum of substitutions (\textit{S}), insertions (\textit{I}), and deletions  (\textit{D}) \citep{woodard1982information}.

\section{Method}
\label{section:method}
A number of active learning approaches are based on the inclusion of \textit{informative} training examples in the dataset for deep learning models, i.e., examples with high error during the training epochs. Such examples have been found to have a greater influence on learning how to correctly label the remaining training data and thus are considered more important than examples with low error (\textit{easier} examples). We first quantify the importance of a training example in the context of a self-supervised ASR system to form a baseline for the comparison of different pruning algorithms.
The training WER of an example after a few training epochs is representative of the difficulty of that example in being transcribed correctly by an ASR system. Intuitively, a hard-to-learn example will have a higher training WER due to the greater misalignment between the generated word sequence and the actual transcription.
We now use the training WER to present three different subset selection strategies for selecting a subset $ {B}_{l}$ of the training data ${D}_{l}$ for fine-tuning a self-supervised speech model on ASR.

\subsection{Strategy 1: Picking the hardest \textit{k} examples}
\label{section:strategy1}

The first approach is to pick the top $k$ training examples, i.e., the ones with the highest WER (Algorithm \ref{alg:topandbottomksubsetselection}). This replicates the pruning strategy of picking the highest error examples \citep{paul2021deep, margatina-etal-2021-active} during training. We first compute the training WER in a particular epoch (WER selection epoch) for all the examples. Then we select examples with the highest WER and perform fine-tuning on this subset. The number of examples selected is determined by the pruning fraction \textit{p}.

\subsection{Strategy 2: Picking the easiest \textit{k} examples} 
\label{section:strategy2}
The second strategy is to pick the bottom $k$ training examples i.e., the ones with the lowest WER (Algorithm \ref{alg:topandbottomksubsetselection}). This is the inverse of strategy \ref{section:strategy1} and removes the harder-to-learn outliers from the training set in an attempt to retain representative examples. 
\begin{algorithm}[!htpb]
   \caption{\textsc Top \textit{k} and Bottom \textit{k} Example Selection for fine-tuning ASR Model}
   \label{alg:topandbottomksubsetselection}
\begin{algorithmic}[1]
   \State {\bfseries Input:} SSL Pretrained Model $f$, Dataset ${D}_{l}$, Strategy $s$ (Top \textit{k} / Bottom \textit{k}), Pruning Fraction $p$, Training Epoch $e$
  \State $W \leftarrow$ Fine-tune $f$ on ${D}_{l}$ and compute WER for each example on epoch $e$
    \State $retainFraction \leftarrow$ $1 - p$
     \State $retainSize \leftarrow retainFraction * len({D}_{l}) $
     \If{$s =  \text{Top \textit{k}}$}
        \State $W \leftarrow sortDescending(W)$ 
    \Else
 \State $W \leftarrow sortAscending(W)$ 
  \EndIf

    \State ${B}_{l} \leftarrow W[0: retainSize]$
\end{algorithmic}
\end{algorithm}

\subsection{Strategy 3: \textsc{Cowerage} Subset Selection} 
\label{section:strategy3}
% \subsection{\textsc{Cowerage} Subset Selection}

\begin{algorithm}[!htpb]
   \caption{\textsc{Cowerage} Subset Selection for fine-tuning ASR Model}
   \label{alg:subsetselection}
\begin{algorithmic}[1]
   \State {\bfseries Input:} SSL Pretrained Model $f$, Dataset ${D}_{l}$, Pruning Fraction $p$, Training Epoch $e$, Bucket Size $b$
   \State $W \leftarrow$ Finetune $f$ on ${D}_{l}$ and compute WER for each example on epoch $e$
    \State $retainFraction \leftarrow$ $1 - p$
   \State ${B}_{l} \leftarrow \emptyset $ 
    \State $W \leftarrow sortDescending(W)$  
    \State $buckets \leftarrow createBuckets(W, size = b)$
  \For{$bucket$ {\bfseries in} $buckets$}
    \State $sampleSize \leftarrow retainFraction*b$  
   \State $S \leftarrow randomSample(bucket, sampleSize)$
   \State ${B}_{l} \leftarrow {B}_{l} \cup S$
  \EndFor
\end{algorithmic}
\end{algorithm}

We now present a novel approach for dataset pruning, which we call \textsc{Cowerage}, i.e., picking examples to ensure the \textit{coverage} of the training WER. The following claim forms the basis of the \textsc{Cowerage} algorithm, which we prove later through multiple experiments (Section \ref{section:empiricaleval}). 

\begin{definition}
\label{claim1}
  Ensuring the coverage of training WER values guarantees the inclusion of phonemically diverse examples in the training data. 
\end{definition}

\begin{figure}[t] 
\begin{center}
\centerline{\includegraphics[width=0.7\columnwidth]{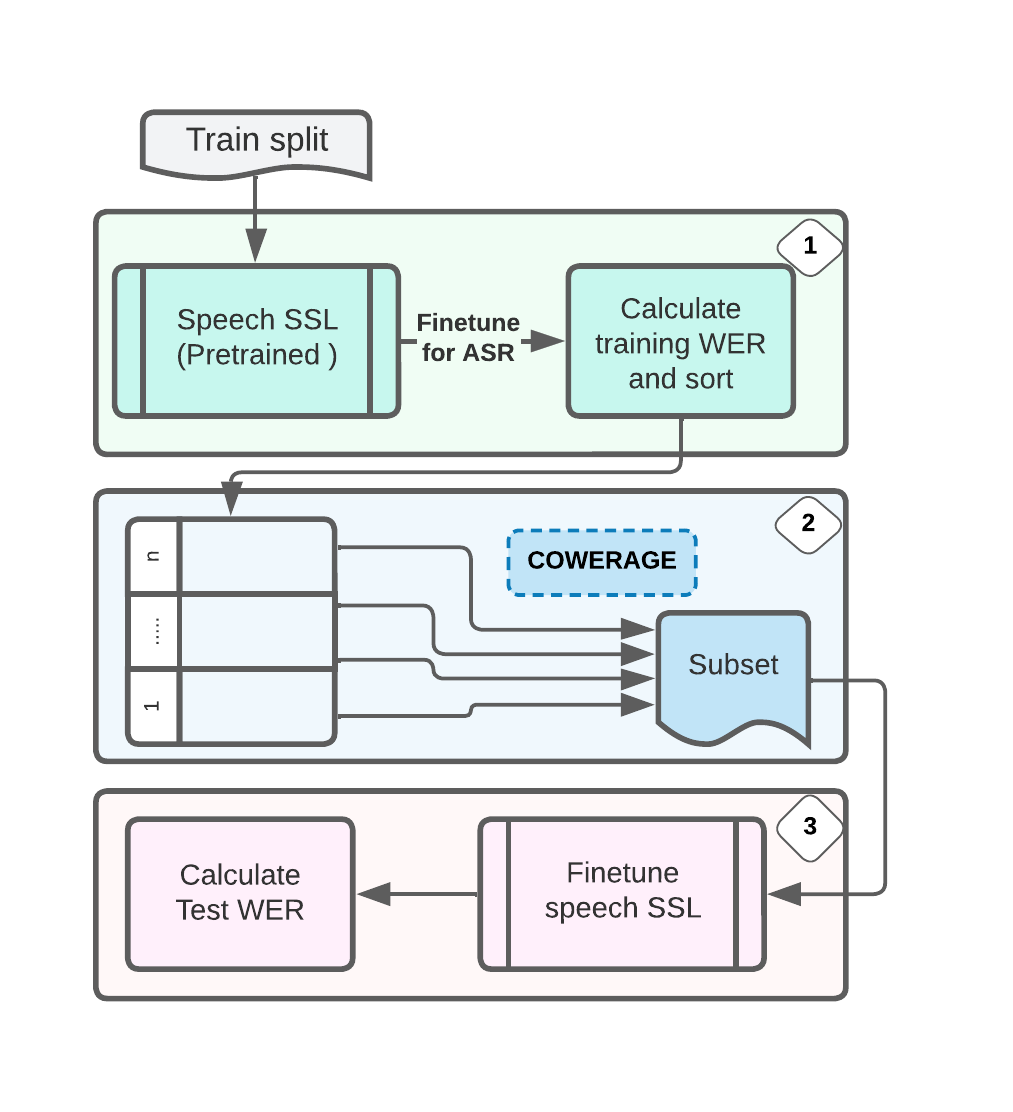}}
\caption{A conceptual representation of the complete flow for fine-tuning a self-supervised model for ASR using a data subset selected by the \textsc{Cowerage} algorithm. In step (1), we perform fine-tuning on the downstream ASR task using the complete dataset and calculate the training WER on a certain epoch. In step (2), we use \textsc{Cowerage} to create a data subset by bucketing the training WER and selecting a subset of examples from each bucket. We then use this data in step (3) to fine-tune the complete model and then evaluate on the test dataset. }
\label{fig:baselineexperiment}
\end{center}
\vskip -0.2in
\end{figure}

With \textsc{Cowerage}, we first compute the training WER for each example in ${D}_{l}$, with the lowest WER as $w_{l}$ and the highest WER as $w_{h}$. We then use a stratified sampling approach of partitioning $N$ total examples from the range $[w_{l}, w_{h}]$ into $M$ buckets, with each bucket defined as,

\begin{equation}
\resizebox{\columnwidth}{!}{%
$S_{i} = \mathcal{W}\left(w_{l}+\frac{i-1}{M}\left(w_{h}-w_{l}\right), w_{l}+\frac{i}{M}\left(w_{h}-w_{l}\right)\right) \;\;\;$%
}
\end{equation}

where $i = 1 \ldots n$. We then use simple random sampling to select $k$ examples uniformly from each bucket,

\begin{equation}
X_{1, \ldots,} X_{k} \sim \mathcal{U}\left(S_{i}\right) 
\end{equation}

where $k$ is decided by the fraction of the dataset to be pruned and the size of the bucket. $\mathcal{U}\left(S_{i}\right)$ denotes the uniform distribution over the set $S_{i}$. This stratified sampling method ensures coverage of WER when selecting training examples. The selected subset is used to fine-tune speech SSL model for ASR and the test performance is evaluated through WER (Fig. \ref{fig:baselineexperiment}). The overall algorithm is presented in Algorithm \ref{alg:subsetselection}.

Our method requires an initial fine-tuning run to compute the ranking of examples, similar to other supervised pruning methods, e.g., EL2N scores \cite{paul2021deep}, RHO-LOSS \cite{mindermann2022prioritized} and Forgetting Norm \cite{azeemi2022dataset}. However, this subset is \textit{transferable} and can subsequently be used for fine-tuning multiple ASR models (Section \ref{section:transferabilityofsubsets}). This amortizes the initial cost of complete training run across the efficiency improvements achieved via multiple fine-tunings done using the created subset. \citet{sorscher2022beyond} identify such pruned datasets as \textit{foundation datasets} which can be used for multiple downstream tasks. 

\subsubsection{Comparison to Random Sampling} 
\label{section:comparisontorandomsampling}
We now highlight the key differences between random subset selection and \textsc{Cowerage}.

\begin{definition}
\label{claim2}
  \textit{In contrast to the \textsc{Cowerage} algorithm, random sampling does not ensure selection of examples from the tail WER range.}
\end{definition}

\begin{proof}
We consider the probability of randomly selecting an example WER ($w$) that is at least at a distance of $k$ standard deviation $\sigma$ from the mean WER. By Chebyshev's inequality: $\operatorname{Pr}(|X-\bar{W}| \geq k \sigma) \leq \frac{1}{k^{2}} = p$, which demonstrates that increasing the WER boundary $w$ (and hence $k$) decreases the probability of randomly selecting a sample with WER greater than $w$.\footnote[1]{Note that the probability of sampling from the tail of the WER degrades \emph{quadratically}.} We now consider the probability of having at least one sample with a WER greater $w$ when we independently draw $n$ samples from the training WER distribution. This is a complement of the event \textit{no sample having a WER greater than $w$ in $n$ draws} which is $(1-p)^n$, and hence the event of interest has the probability upper bound $1-(1-p)^n = 1-(1-\frac{1}{k^2})^n$. This demonstrates that decreasing the sample size and increasing the pruning percentage reduces the probability of selecting a tail WER example. In contrast, for \textsc{Cowerage}, the probability of selecting at least one example with a WER greater than $ \bar{W} + k\sigma$ is $Pr(|S_{i}| > 0) = q$, where $S_i$ is a tail bucket with the WER range $(a, b)$ such that $a \geq \bar{W} + k\sigma$ and $b>a$. This probability ($q$) approaches 1 if we consider a bucket size satisfying the range $(a, b)$, and hence \textsc{Cowerage} ensures selection of examples from the tail WER range.

\vskip -0.2in
\end{proof}

\begin{definition}
\label{claim3}
  \textit{Subsets selected by \textsc{Cowerage} have a lower variance of the sample mean of WER than randomly selected samples.}
\end{definition}

\begin{proof}   We first consider the variance of samples selected by \textsc{Cowerage}. Let $S_{ij}$ be the sample $i$ from bucket $S_j$. The average WER in bucket $j$ is  $\bar{W}_{j} = \frac{\sum_{i} S_{ij}}{k}$, variance in bucket $j$ is $\sigma_{j}^{2}$ and the overall average is $\bar{W} = \frac{\sum_{j} \bar{W}_{j}}{M}$. The variance of the sample mean of WER is,

\begin{equation}
\label{eq:stratified}
\mathrm{Var}_\textsc{cowerage}[\bar{W}]=\frac{\sum_{j} \mathrm{Var}\left[\bar{W}_{j}\right]}{M^{2}}
\end{equation} 

$\mathrm{Var}\left[\bar{W}_{j}\right]$ is the variance of the sample mean within a particular bucket and is equivalent to $\frac{\sigma_{j}^{2}}{k}$. Thus, we get
\begin{equation}
\resizebox{0.9\columnwidth}{!}{
\label{eq:stratified}
 $\mathrm{Var}_\textsc{cowerage}[\bar{W}]=\frac{\sum_{j} \mathrm{Var}\left[\bar{W}_{j}\right]}{M^{2}} =\frac{ \sum_{j} \sigma_{j}^{2}}{M^{2}k} =\frac{ \sum_{j} \sigma_{j}^{2}}{MN}$ }
\end{equation}

Now we consider the variance of a simple random sample. $\mathrm{Var}[\bar{W}]=\frac{\sigma^2}{N} $ with $\sigma^{2} =\mathbb{E}\left[W^{2}\right]-\mu^{2}$. Considering the contribution from each bucket in the random sample, we can specify $\sigma^2 =  \frac{ \sum_{j} \mathbb{E}\left[S_{j}\right]}{M}-\mu^{2} =\frac{  \sum_{j}\left(\mu_{j}^{2}+\sigma_{j}^{2}\right)}{M} -\mu^{2}  =\frac{ \sum_{j}\left(\left(\mu_{j}-\mu\right)^{2}+\sigma_{j}^{2}\right)}{M}  $. Thus,
\begin{equation}
\label{eq:random}
\mathrm{Var_\textsc{random}}[\bar{W}] = \frac{ \sum_{j}\left(\left(\mu_{j}-\mu\right)^{2}+\sigma_{j}^{2}\right)}{MN}
\end{equation}

Comparing \eqref{eq:stratified} and \eqref{eq:random}, $\mathrm{Var_\textsc{random}}[\bar{W}] \geq \mathrm{Var_\textsc{cowerage}}[\bar{W}] $ and the result follows.

\vskip -0.2in
\end{proof}

%\begin{figure}[!ht]

\begin{table*}[!htbp]
\centering
\caption{Test WER for the four strategies of pruning the training set evaluated at multiple pruning fractions (0.1, 0.3, 0.5, 0.7, 0.9) and different datasets (LJSpeech, LS-10h and TIMIT). The training WER in a particular epoch is averaged over 10 runs and then used for a particular pruning strategy. For each result, we do three independent runs and report the mean test WER. The standard deviation is reported in Table \ref{tab:pruningresultsstd}. \textsc{Cowerage} consistently demonstrates the lowest WER at various pruning fractions. WER selection epoch (WSE) is set to 8 for these experiments. See Section \ref{section:werselectionepoch} for WSE ablation. }

\resizebox{\textwidth}{!}{%

\begin{tabular}{c  c  c c c c c c c c c c  c c c c c c  c c c c c }
%\small
\toprule
\textbf{Dataset} &  \textbf{Strategy} & \multicolumn{6}{c}{\texttt{wav2vec2-base}} & \multicolumn{6}{c}{\texttt{HuBERT-base}}  \\
\cmidrule(lr){3-8}
\cmidrule(lr){9-14}

{} & {} & \textbf{No pruning} & \textbf{0.1} & \textbf{0.3} & \textbf{0.5} & \textbf{0.7} & \textbf{0.9} & \textbf{No pruning} & \textbf{0.1} & \textbf{0.3} & \textbf{0.5} & \textbf{0.7} & \textbf{0.9} \\

\midrule
LJSpeech   &  Random  & 0.052 & 0.062   & 0.071 		 & 0.085 		 & 0.128		& 0.251 & 0.091 & 0.117 & 0.128 & 0.140 & 0.196 & 0.272 \\
{}   &  Top K         & 0.052 & 0.060   & 0.064		     & 0.077 	     & 0.101	    & 0.238	 & 0.091 & 0.109 & 0.118 & 0.135 & 0.168 & 0.248 \\
{}   &  Bottom K & 0.052 & 0.057    & 0.063	  & 0.070 	 & 0.091  &	  0.166  & 0.091 & 0.105 & 0.116 & 0.130 & 0.151 & 0.181 \\
{}   &  \textsc{Cowerage} & \textbf{0.052} & \textbf{0.054} &	\textbf{0.060}  &	\textbf{0.067}	  &	\textbf{0.085}  &	\textbf{0.144} & \textbf{0.091}   & \textbf{0.101} & \textbf{0.107} & \textbf{0.115} & \textbf{0.136} & \textbf{0.153}	 \\

\midrule
LS-10h   &  Random   & 0.140 & 0.147    &	0.168  & 0.188	 	 &	0.245 & 0.360 & 0.180 & 0.219 & 0.220 & 0.298 & 0.309 & 0.424 \\
{}   &  Top K   & 0.140 & 0.143   & 0.155		 & 0.174	 	 &	0.198		 & 0.343 & 0.180 & 0.210 & 0.215 & 0.268 & 0.313 & 0.391	 \\
{}   &  Bottom K  & 0.140 & 0.146   & 0.159	 	 &  0.175	 &0.201	 & 0.336 & 0.180 & 0.215 & 0.219  & 0.269 & 0.336 & 0.381 \\
{}   &  \textsc{Cowerage} & \textbf{0.140} & \textbf{0.142} & 	\textbf{0.150} &		\textbf{0.164}	  &	\textbf{0.192}  &	\textbf{0.277} & \textbf{0.180} & \textbf{0.185} & \textbf{0.211} 	& \textbf{0.250} & \textbf{0.290} & \textbf{0.341} \\

\midrule
TIMIT    &  Random  & 0.315 &  0.325 & 0.341 & 0.357 & 0.394 & 0.557 & 0.328 & 0.357 & 0.373 & 0.392 & 0.452 & 0.675\\
{}   &  Top K  & 0.315 & 0.322 &		0.334 &		0.392 &	0.472  &	0.678  & 0.328 & 0.345  &  0.366  &  0.435  &  0.532  &  0.871\\
{}   &  Bottom K & 0.315 & 0.336  &	0.360  &	0.411  &	0.521 & 	0.887 & 0.328 & 0.346 & 0.391 & 0.447 & 0.568 & 0.931 \\
{}   &  \textsc{Cowerage}  & \textbf{0.315} & \textbf{0.320}   &	 \textbf{0.333}  &	\textbf{0.339}		& \textbf{0.369} &	\textbf{0.455} & \textbf{0.328}  & \textbf{0.335} & \textbf{0.355} & \textbf{0.381} & \textbf{0.445} & \textbf{0.616} \\

\bottomrule
\end{tabular}}

\label{tab:pruningresults}
\end{table*}

\section{Configurations}

\subsection{Models} 
We use the \texttt{wav2vec2-base} \cite{baevski2020wav2vec} (95M parameters) and \texttt{HuBERT-base} model \cite{hsu2021hubert} (90M parameters) for our experiments. \texttt{wav2vec2} consists of a CNN-based encoder that processes the input waveform which is then discretized via the quantization layer and passed to the BERT module where the actual contextual representation is learned. \texttt{HuBERT} learns a combined language and acoustic model through a prediction loss which is applied to masked regions only. We select \texttt{wav2vec2-base} and \texttt{HuBERT-base} that is pre-trained on Librispeech 960h, and fine-tune them for ASR using the Connectionist Temporal Classification (CTC) loss \citep{graves2006connectionist} on the subsets of three speech datasets: TIMIT \citep{garofolo1993darpa}, Librispeech 10h \citep{panayotov2015librispeech} and LJSpeech \citep{ito2017lj}. We report WER for pruning fractions of 0.1, 0.3, 0.5, 0.7, and 0.9 to adequately evaluate low, moderate, and extreme pruning settings across different strategies. Please see 
Appendix \ref{section:dataappendix} for details about train and test splits and Appendix \ref{section:trainingdetails} for hyperparameters.

\subsection{Baseline} We consider the baseline experiment of randomly pruning the train split of the dataset on multiple fractions and fine-tuning the ASR model on the generated subset. The performance evaluation is done through WER on the test set.

\section{Empirical Evaluation}
\label{section:empiricaleval}
\textbf{Experiments.} We fine-tune \texttt{wav2vec2-base} model on the selected dataset and calculate the WER of the training examples over ten independent runs. The training scores (averaged over 10 runs) from a particular epoch are then used to prune the examples through the pruning strategies (\ref{section:strategy1}, \ref{section:strategy2}, \ref{section:strategy3}) to generate a subset of training data.  The data subsets are then used to fine-tune \texttt{wav2vec2-base} and \texttt{HuBERT-base} for ASR. The training WER distribution and the subsets of TIMIT, Librispeech and LJSpeech selected through each method are shown in Appendix \ref{section:trainingwerdistribution}.

% .

\textbf{Results.} We show the results of pruning experiments via different strategies across multiple pruning fractions in Table \ref{tab:pruningresults}. For each strategy and pruning fraction, we report the mean WER of three independent runs. The variability across runs is shown in Appendix \ref{tab:pruningresultsstd}.  We observe that for the majority of pruning fractions, \textsc{Cowerage} subset selection is consistently better than the other three pruning strategies (top \textit{k}, bottom \textit{k}, and random pruning) for TIMIT, LS-10h, and LJSpeech. At higher pruning fractions, the difference between the test WER for \textsc{Cowerage} and the other pruning strategies increases, e.g., on the Librispeech-10h dataset with 90\% pruning, \textsc{Cowerage} shows 17\% relative WER improvement over Bottom K strategy compared to 5\% relative WER improvement at 30\% pruning. This observation can also be made for random sampling and is consistent with claim \ref{claim2} where we consider the impact of smaller sample sizes (higher pruning percentages) on the selection of examples from tail WER which subsequently affects test error. On the TIMIT dataset, going from 10\% pruning to 90\% pruning leads to an absolute increase of only 0.135 WER for \textsc{Cowerage}  compared to an increase of 0.551, 0.356, and 0.232 for Bottom K, Top K, and Random respectively.

\subsection{Transferability of representative subsets} 
\label{section:transferabilityofsubsets}
Table \ref{tab:pruningresults} shows that \textsc{Cowerage} demonstrates better performance in the fine-tuning run of \texttt{HuBERT-base} on the subsets constructed through training WER values of \texttt{wav2vec2-base}. The relative trend for other pruning strategies is also similar to that of \texttt{wav2vec2-base}. This suggests that the representative subsets computed through one speech SSL model are \textit{transferable} to another speech SSL model, making them \textit{model-agnostic} and \textit{dataset-specific}. We also verify this transferability for a larger model (\texttt{wav2vec2-large}) and the results are shown in Appendix \ref{section:largermodels}. This property is present in a few other pruning metrics for deep learning models as well, including EL2N score \cite{paul2021deep} and RHO-loss \cite{mindermann2022prioritized}. Our explanation is that since the composition of the representative subset is more influenced by the \textit{ranking} of training examples instead of absolute WER values (line 5-6 of Algorithm \ref{alg:subsetselection}), it makes them relevant for fine-tuning other speech SSL models. Additionally, the prior averaging of the training WER values theoretically eliminates the influence of specific model weights, which produces a more precise ranking of the examples. We can consider the representative subsets constructed through \textsc{Cowerage} as \textit{foundation datasets} \cite{sorscher2022beyond} which need to be constructed once and can be later used to fine-tune multiple other speech SSL models.

% \clearpage

\subsection{Ablation study}

\textbf{The Impact of Offset.} To identify whether there is another contiguous subset of examples below the ones with the highest WER which can perform better than random pruning, we introduce an offset while selecting the top \textit{k} training examples, mirroring the protocol presented by \citet{paul2021deep}. We compute the training WER for the examples and sort them in ascending order. We then maintain a sliding window from offset $k$ to $k + N$ which keeps $N$ data points but incrementally excludes the training examples with the highest WER. For offset sizes from 0 to 500, we notice a change in the test WER but no single offset size is consistently better than random pruning. An important implication of this finding is that no contiguous subset of training examples picked according to the WER is better than random pruning in the TIMIT speech corpus, contrary to the previous studies on vision datasets that have shown a clear correlation between the top-scoring examples and the accuracy \citep{paul2021deep}.

\begin{figure}[t]
    \centering
    \includegraphics[width = 0.8\columnwidth]{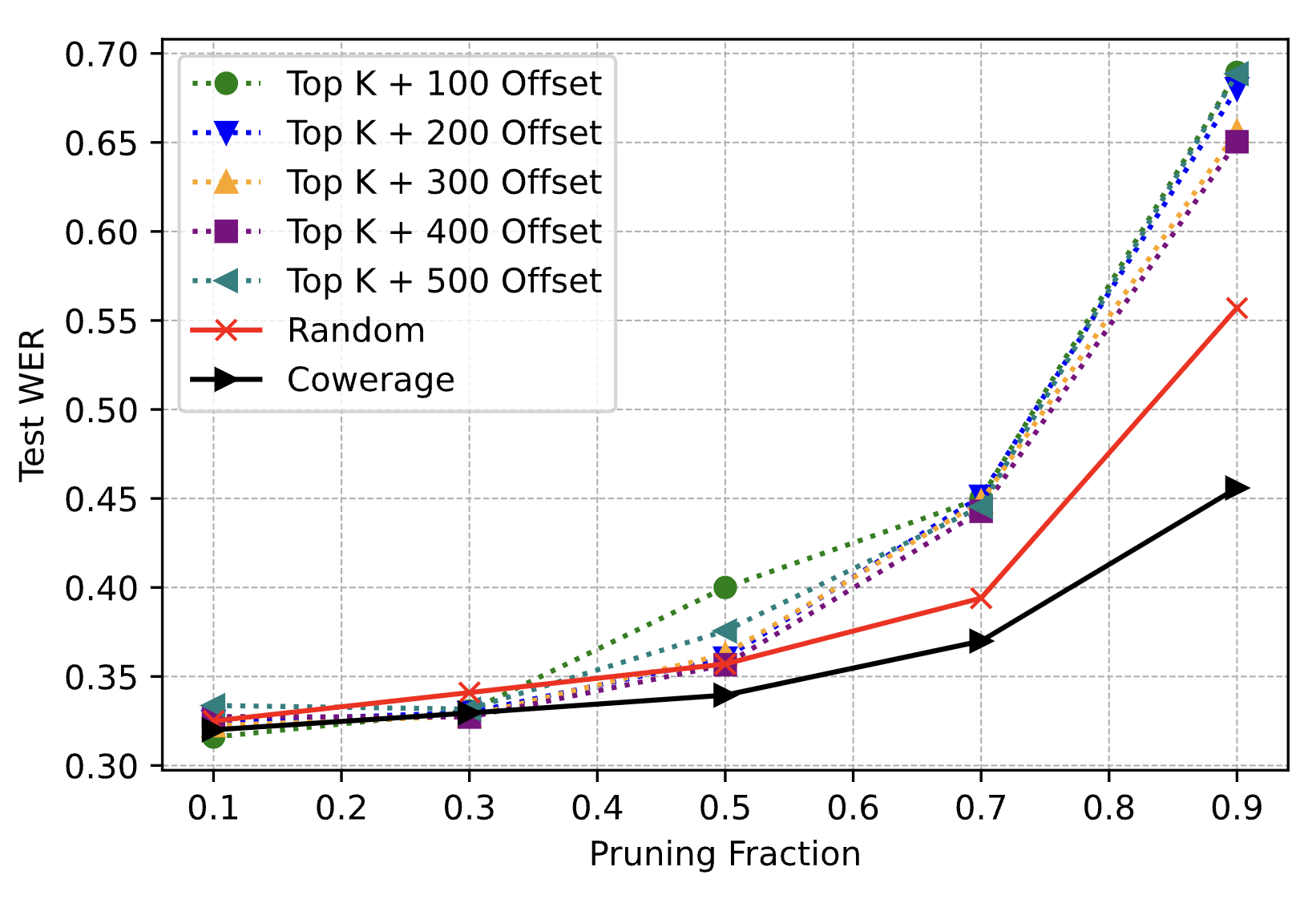}
    \caption{The test WER for the different offsets while picking the top \textit{k} examples compared over different pruning fractions of the TIMIT dataset. Note that no single offset consistently performs better than random pruning.}
    \label{differentoffsets}    
    \vskip -0.2in
\end{figure}

\textbf{Selection within the buckets.}
 The strategy proposed in the original \textsc{Cowerage} algorithm is to randomly sample elements from each bucket. We also evaluate two other strategies: picking the first \textit{k} examples within each bucket and picking the last \textit{k} ones, similar to strategies 1 and 2 except that now we are sampling within a particular bucket. The results in Table \ref{tab:withinbucketselection} show that the random selection outperforms other strategies. Additionally, we evaluate the impact of increasing the bucket size on the test WER in Appendix \ref{section:selectingbuckets}.

\begin{table}[!htb]
\caption{Test WER for different strategies of picking samples within each bucket for \textsc{Cowerage} algorithm on 0.7 pruning fraction and WER Selection Epoch 8.}
\label{tab:withinbucketselection}
\begin{center}
\begin{small}

\resizebox{1\columnwidth}{!}{%
\begin{tabular}{lccr}
\toprule
 & \textsc{Cowerage} + Top \textit{k} & \textsc{Cowerage} + Bottom \textit{k} & \textsc{Cowerage} + Random \\
\midrule
WER & $0.378 \pm 0.002$   & $0.401 \pm 0.002$ & \textbf{$0.369 \pm 0.004$ }  \\
% Test WER    & 0.487 & 0.493 & \textbf{0.474 }  \\
\bottomrule

\end{tabular}}
\end{small}
\end{center}

\end{table}

\subsection{Phoneme Recognition on TIMIT}
\label{section:phonemerecognitionontimit}

We evaluate the subset selection methods on the task of phoneme recognition with \texttt{wav2vec2-base} on TIMIT dataset and report the phoneme error rate (PER) on the test set (Table \ref{tab:phonemerecognition}). \textsc{Cowerage} consistently demonstrates the lowest PER on all the pruning fractions above 0.2.

\begin{table}[!htbp]
\centering
\caption{Phoneme recognition on the TIMIT dataset with \texttt{wav2vec2-base}. We report PER for multiple pruning fractions and different strategies.}
\vskip 0.1in

\resizebox{0.8\columnwidth}{!}{%
\begin{tabular}{c c c c c c c}
%\small
\toprule
Strategy & \multicolumn{5}{c}{Pruning Fraction}\\
\midrule
\midrule
{} & \textbf{0.1} & \textbf{0.3} & \textbf{0.5}  & \textbf{0.7}  & \textbf{0.9} \\
\midrule
Random  &  0.124 & 	0.133 &		0.148 &		0.230  &	1.000\\
Top K  & \textbf{0.118} &  0.137 & 0.168 & 	0.244 &		1.000\\
Bottom K & 0.122  &	0.142  &	0.170  &	0.282 &	1.000 \\
\textsc{Cowerage} & 0.120  &	\textbf{0.133}  &	\textbf{0.145}		& \textbf{0.211}  &	1.000\\

\bottomrule
\end{tabular}}
\label{tab:phonemerecognition}
\end{table}

\subsection{Training time for subsets}
\label{section:trainingtimeexperiment}
Practically, the choice of pruning fraction can be made according to the intended size of the final dataset under the given time and memory constraints. We conduct an experiment to determine the total steps required for convergence and the real training time for \texttt{wav2vec2} on TIMIT. The results are shown in Table \ref{tab:trainingtime} (for a constant learning rate). We report the real training time for the pruned datasets as a fraction of the training time for the complete dataset ($x$) for relative comparison. There is a significant reduction in training time for higher pruning fractions.

    \begin{table}[!hptb]
    \centering
    \caption{Steps required for convergence and training time for \texttt{wav2vec2} on TIMIT for different pruning fractions. We replicate the results of \textsc{Cowerage} from Table \ref{tab:pruningresults} for relative comparison.}

    \resizebox{\columnwidth}{!}{%
   \begin{tabular}{c c c c c c c} 
   \toprule
Pruning Fraction & 0.9 & 0.7 & 0.5 & 0.3 & 0.1 & 0 \\
\midrule
Steps required for convergence & 1050 & 1900 & 2400 & 2800 & 3170 & 3350  \\
\midrule
Training time & $0.42\times$ & $0.62\times$ & $0.77\times$ & $0.85\times$ & $0.90\times$ & $\times$ \\
\midrule
Test WER (\textsc{Cowerage}) & 0.455 & 0.369 & 0.339 & 0.333 & 0.320 & 0.315 \\

 \bottomrule
\end{tabular}}

    \label{tab:trainingtime}
    \end{table}

\section{Connection to Phonemes}
\label{section:connectiontophonemes}

To understand why \textsc{Cowerage} performs better than other pruning strategies, it is important to find out how does the phoneme distribution of training examples vary with the training error during fine-tuning of the self-supervised speech recognition models. We now perform empirical analysis to verify claim \ref{claim1}. For this analysis, we select the standard TIMIT dataset as it contains time-aligned, hand-verified phonetic and word transcriptions for each training example.

\begin{figure}[t]
    \centering
    \includegraphics[width = 0.8\columnwidth]{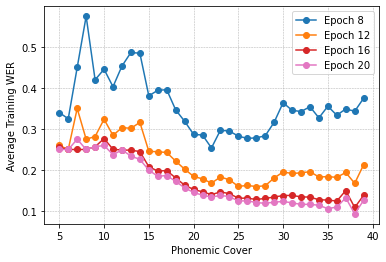}
    \caption{The training WER and the phonemic cover of examples in TIMIT dataset (without pruning) compared over multiple training epochs. The WER is computed by averaging the training scores of the examples with the same phonemic cover. The training scores for each training example and a particular epoch are computed by averaging over 10 runs.}
    \label{fig:trainingwervsphonemiccover} 
    \vskip -0.1in

\end{figure}

We first record the training WER of each training example in the TIMIT dataset over 10 runs and average it. Then, we compute the total number of unique phonemes in each example, which we call the \textit{phonemic cover}. Subsequently, we group together the training examples with same phonemic cover and calculate the average training WER for each group (Fig. \ref{fig:trainingwervsphonemiccover}). In the earlier training epochs, the examples with a relatively low ($<17$) or a high ($>28$) phonemic cover have a greater WER (blue line in Fig. \ref{fig:trainingwervsphonemiccover}) as compared to the examples with a moderate number of phonemes ($17\leq \texttt{phonemicCover}\leq 28$). In the later epochs ($\geq12$), the inverse relationship between the training WER and the phonemic cover becomes more evident; the examples with a greater number of distinct phonemes have a lower training WER and vice versa.

\textbf{Significance.} This relationship between the training WER and the phonemic cover has several implications. 
Firstly, it demonstrates that there is a sizable population of sentences with a low phonemic cover that are harder to learn and hence represent a high training WER.  Similarly, there are many low WER sentences with a high phonemic cover (examples are presented in Appendix \ref{trainandtestexamples}). More importantly, this experiment validates our claim that ensuring the coverage of training WER values in a particular subset leads to the inclusion of phonemically diverse training examples \textit{without} explicitly learning any phoneme-level error model. This is beneficial as accurate phonetic data is not available for the majority of 7000 spoken languages \citep{billington21_interspeech}. In contrast, any method that directly ensures phoneme diversity requires an accurate phonetic transcription beforehand, which is a resource-intensive process requiring manual labeling by linguists.

% \subsection{Statistical differences in phoneme distribution}
% \label{section:statisticaldifferences}
To verify if the difference between the phoneme distributions of the examples within the \textsc{Cowerage} subset and the other two strategies (top \textit{k} and bottom \textit{k}) is statistically significant, we conduct the Mann-Whitney U test, a non-parametric test, at a significance level of 0.01. 
We found that the differences were statistically significant at the 1\% level (\textit{p}-value $< 0.01$).
%The differences are found to be statistically significant, with a \textit{p}-value $< .01$.
The results are shown in Table \ref{tab:statisticaldifference}.

\begin{table}[!hptb]

\centering
\caption{The statistical significance of the difference between the phoneme distribution of the examples within the \textsc{Cowerage} subset and the other two strategies (top \textit{k} and bottom \textit{k}). MWU: Mann-Whitney U.}

\resizebox{0.8\columnwidth}{!}{%
\begin{tabular}{ c c c }
\toprule
{} &  MWU & \textit{p}-value\\
\midrule

Top \textit{k} vs \textsc{Cowerage} & 2146027.5 & $< 0.001$ \\
\midrule
Bottom \textit{k} vs \textsc{Cowerage} & 2229653.0 & $< 0.001$ \\
\bottomrule
\end{tabular}}
\label{tab:statisticaldifference}
\end{table}

\subsection{Phonemic diversity and latent representation in speech SSL}
\label{section:latentrepresentation}
How does phonemic diversity impact the discrete latent speech representations within self-supervised speech recognition models? To answer this, we study the latent representation ($\mathbf{q}_{t}$) learned by the quantizer within \texttt{wav2vec2} for different phonemes. \citet{baevski2020wav2vec} analyze the conditional probability $P\left(phoneme  \mid \mathbf{q}_{t}\right)$ for each of the 39 phonemes in the TIMIT train set by computing the co-occurence between the phonemes and speech latents (see Appendix D of \citet{baevski2020wav2vec}). They demonstrate that different discrete latents specialize in different phonetic sounds in \texttt{wav2vec2} model. Building upon this, \citet{shim2021understanding} analyze the relationship between attention and phonemes in Transformer-based ASR models by considering the attention map that extracts phonologically meaningful features. They observe that the characteristic feature of phonetic localization is the higher attention weights assigned to similar phonemes in the attention map (see Fig. 3 of \citet{shim2021understanding}). Given these observations, we hypothesize that the performance gains for \textsc{Cowerage} are due to the greater phonemic diversity which enables a more robust latent representation of each phoneme in \texttt{wav2vec2}. This view is supported by the results in Table \ref{tab:pruningresults} which demonstrate bigger gains in test WER for higher pruning fractions in \textsc{Cowerage}. We conjecture that this is due to greater example diversity provided by \textsc{Cowerage} and lack of representation of examples from the tail WER range in the case of other approaches.

\section{Related Work}

\textbf{Data pruning.} Devising strategies for data pruning and constructing optimal subsets is a recent topic of interest in the area of optimization, coresets and efficient deep learning \citep{ tolochinsky2018coresets, dong2019speech, mirzasoleiman2020coresets, huang2021coresets, jiang2021coresets, jubran2021coresets, durga2021training, kothawade2021similar, killamsetty2021grad, kothyari2021personalizing, azeemi2022dataset}. A few studies have examined the training landscape for drawing clues about the optimal subset creation \citep{toneva2018empirical, agarwal2020estimating,  baldock2021deep, paul2021deep, schirrmeister2022less}.  \citet{paul2021deep} evaluate the impact of static data pruning on the performance on standard vision datasets (e.g., CIFAR-10 and CIFAR-100) and models (ResNet). They use the gradient norm (GraNd) and the error norm (EL2N) for removing the \textit{easy} training examples and pruning a significant chunk of the dataset without affecting the generalization error. The authors observe that the local information in the early training epochs is a strong indicator of the importance of training examples and thus can be used to select a good subset of training data effectively. This is consistent with our observation regarding WER selection epoch.

\textbf{Selection of hard-to-learn examples.} Although sampling hard-to-learn examples has been a popular choice for data pruning in deep learning models, it appears to work on a limited set of tasks that share certain properties. A study on visual question answering (VQA) \citep{karamcheti2021mind} demonstrates that the active learning approaches that prefer picking the \textit{harder} examples do not outperform random pruning on the VQA task across multiple models and datasets. The authors demonstrate the role of collective outliers \citep{han2011data} in degrading the generalization performance and find out that the preference for selecting these harder-to-learn outliers by the active learning methods is the cause of poor improvements in efficiency as compared to random sampling. Our findings regarding harder-to-learn training examples are similar; they do not consistently perform better for speech SSL models.

\textbf{Data subset selection for ASR.} The existing work on active learning and data pruning for ASR systems emphasize the importance of ensuring phonemically rich text and higher coverage of words \citep{wu2007data, ni2015unsupervised, wei2014submodular, mendoncca2014method, ni2015submodular, ni2015unsupervised, ni2016cross}. An early study \citep{wu2007data} demonstrates that selecting a subset that is sampled uniformly across phonemes and words is more effective than random sampling. A subsequent work \citep{wei2014submodular} proposes a method for selecting the data by maximizing a constrained sub-modular function. The results show the possibility of a significant reduction of the training data when using acoustic models based on Gaussian mixture models. 

In ASR, active learning aims to select the most informative utterances to be transcribed from a large amount of un-transcribed utterances. In contrast, our core objective is to construct an optimal data subset by selecting the informative and representative examples from a \textit{fully labeled} dataset i.e. the examples for which audios and the reference transcriptions are available.

The majority of these existing approaches have focused on the earlier ASR systems instead of the Deep Neural Network (DNN) based models. Although model pruning has been explored for self-supervised and other ASR models \citep{lai2021parp, wu2021dynamic, zhen2021sparsification}, data subset selection for fine-tuning self-supervised ASR systems has only been explored in the context of personalization for accented speakers \citep{awasthi2021error}. A phoneme-level error model is proposed which selects sentences that yield a lower test WER as compared to random sentence selection. In contrast, our \textsc{Cowerage} algorithm has the advantage that no complex, dataset-specific phoneme-level error model needs to be learned, which constructs phonemically diverse subsets. Instead, just the training WER can be used to devise a strategy for pruning that performs better than random selection. Additionally, to the best of our knowledge, this is the first study considering data subset selection for efficient fine-tuning in self-supervised speech recognition models.

\section{Conclusion and Future Work}
\label{section:limitationsandfuturework} 
In this work, we proposed \textsc{Cowerage}, a new method for pruning data for self-supervised automatic speech recognition, which relies on sampling data in a way that ensures coverage of training WER values. 
An evaluation on \texttt{wav2vec2} and \texttt{HuBERT} and three datasets show that \textsc{Cowerage} performs better than random selection and other data pruning strategies that select harder-to-learn or easier-to-learn examples. We demonstrate that the pruned subsets are transferable to other speech SSL models which amortizes the cost of initial training run across the efficiency improvements achieved via multiple fine-tunings done using the created subset. We unveil the connection between the training word error rate and the phonemic cover of training examples across multiple training epochs and analyze the pruning results through this lens. We show that \textsc{Cowerage} outperforms other subset selection strategies as it ensures phonemic diversity within the training examples by directly utilizing the training WER of speech SSL models. While we designed our approach to be dataset agnostic and applicable to different distributions of training WER, it remains to be empirically evaluated whether our methodology generalizes to noisier data and multilingual speech corpora.

\bibliographystyle{icml2022}
\bibliography{example_paper}

\begin{thebibliography}{46}
\providecommand{\natexlab}[1]{#1}
\providecommand{\url}[1]{\texttt{#1}}
\expandafter\ifx\csname urlstyle\endcsname\relax
  \providecommand{\doi}[1]{doi: #1}\else
  \providecommand{\doi}{doi: \begingroup \urlstyle{rm}\Url}\fi

\bibitem[Agarwal et~al.(2020)Agarwal, D'souza, and
  Hooker]{agarwal2020estimating}
Agarwal, C., D'souza, D., and Hooker, S.
\newblock Estimating example difficulty using variance of gradients.
\newblock \emph{arXiv preprint arXiv:2008.11600}, 2020.

\bibitem[Ahmed \& Wahed(2020)Ahmed and Wahed]{ahmed2020democratization}
Ahmed, N. and Wahed, M.
\newblock The de-democratization of ai: Deep learning and the compute divide in
  artificial intelligence research.
\newblock \emph{arXiv preprint arXiv:2010.15581}, 2020.

\bibitem[Awasthi et~al.(2021)Awasthi, Kansal, Sarawagi, and
  Jyothi]{awasthi2021error}
Awasthi, A., Kansal, A., Sarawagi, S., and Jyothi, P.
\newblock Error-driven fixed-budget asr personalization for accented speakers.
\newblock In \emph{ICASSP 2021-2021 IEEE International Conference on Acoustics,
  Speech and Signal Processing (ICASSP)}, pp.\  7033--7037. IEEE, 2021.

\bibitem[Azeemi et~al.(2022)Azeemi, Qazi, and Raza]{azeemi2022dataset}
Azeemi, A.~H., Qazi, I.~A., and Raza, A.~A.
\newblock Dataset pruning for resource-constrained spoofed audio detection.
\newblock \emph{Proc. Interspeech 2022}, pp.\  416--420, 2022.

\bibitem[Baevski et~al.(2020)Baevski, Zhou, Mohamed, and
  Auli]{baevski2020wav2vec}
Baevski, A., Zhou, H., Mohamed, A., and Auli, M.
\newblock wav2vec 2.0: A framework for self-supervised learning of speech
  representations.
\newblock \emph{arXiv preprint arXiv:2006.11477}, 2020.

\bibitem[Baldock et~al.(2021)Baldock, Maennel, and Neyshabur]{baldock2021deep}
Baldock, R.~J., Maennel, H., and Neyshabur, B.
\newblock Deep learning through the lens of example difficulty.
\newblock \emph{arXiv preprint arXiv:2106.09647}, 2021.

\bibitem[Billington et~al.(2021)Billington, Stoakes, and
  Thieberger]{billington21_interspeech}
Billington, R., Stoakes, H., and Thieberger, N.
\newblock {The Pacific Expansion: Optimizing Phonetic Transcription of Archival
  Corpora}.
\newblock In \emph{Proc. Interspeech 2021}, pp.\  4029--4033, 2021.
\newblock \doi{10.21437/Interspeech.2021-2167}.

\bibitem[Coleman et~al.(2019)Coleman, Yeh, Mussmann, Mirzasoleiman, Bailis,
  Liang, Leskovec, and Zaharia]{coleman2019selection}
Coleman, C., Yeh, C., Mussmann, S., Mirzasoleiman, B., Bailis, P., Liang, P.,
  Leskovec, J., and Zaharia, M.
\newblock Selection via proxy: Efficient data selection for deep learning.
\newblock \emph{arXiv preprint arXiv:1906.11829}, 2019.

\bibitem[Dong et~al.(2019)Dong, Guo, and Wu]{dong2019speech}
Dong, L., Guo, Q., and Wu, W.
\newblock Speech corpora subset selection based on time-continuous utterances
  features.
\newblock \emph{Journal of Combinatorial Optimization}, 37\penalty0
  (4):\penalty0 1237--1248, 2019.

\bibitem[Durga et~al.(2021)Durga, Iyer, Ramakrishnan, and
  De]{durga2021training}
Durga, S., Iyer, R., Ramakrishnan, G., and De, A.
\newblock Training data subset selection for regression with controlled
  generalization error.
\newblock In \emph{International Conference on Machine Learning}, pp.\
  9202--9212. PMLR, 2021.

\bibitem[Garofolo et~al.(1993)Garofolo, Lamel, Fisher, Fiscus, and
  Pallett]{garofolo1993darpa}
Garofolo, J.~S., Lamel, L.~F., Fisher, W.~M., Fiscus, J.~G., and Pallett, D.~S.
\newblock Darpa timit acoustic-phonetic continous speech corpus cd-rom. nist
  speech disc 1-1.1.
\newblock \emph{NASA STI/Recon technical report n}, 93:\penalty0 27403, 1993.

\bibitem[Graves et~al.(2006)Graves, Fern{\'a}ndez, Gomez, and
  Schmidhuber]{graves2006connectionist}
Graves, A., Fern{\'a}ndez, S., Gomez, F., and Schmidhuber, J.
\newblock Connectionist temporal classification: labelling unsegmented sequence
  data with recurrent neural networks.
\newblock In \emph{Proceedings of the 23rd international conference on Machine
  learning}, pp.\  369--376, 2006.

\bibitem[Han et~al.(2011)Han, Pei, and Kamber]{han2011data}
Han, J., Pei, J., and Kamber, M.
\newblock \emph{Data mining: concepts and techniques}.
\newblock Elsevier, 2011.

\bibitem[Hsu et~al.(2021)Hsu, Bolte, Tsai, Lakhotia, Salakhutdinov, and
  Mohamed]{hsu2021hubert}
Hsu, W.-N., Bolte, B., Tsai, Y.-H.~H., Lakhotia, K., Salakhutdinov, R., and
  Mohamed, A.
\newblock Hubert: Self-supervised speech representation learning by masked
  prediction of hidden units.
\newblock \emph{arXiv preprint arXiv:2106.07447}, 2021.

\bibitem[Huang et~al.(2021)Huang, Sudhir, and Vishnoi]{huang2021coresets}
Huang, L., Sudhir, K., and Vishnoi, N.
\newblock Coresets for time series clustering.
\newblock \emph{Advances in Neural Information Processing Systems}, 34, 2021.

\bibitem[Ito \& Johnson(2017)Ito and Johnson]{ito2017lj}
Ito, K. and Johnson, L.
\newblock The lj speech dataset, 2017.

\bibitem[Jiang et~al.(2021)Jiang, Krauthgamer, Wu, et~al.]{jiang2021coresets}
Jiang, S., Krauthgamer, R., Wu, X., et~al.
\newblock Coresets for clustering with missing values.
\newblock \emph{Advances in Neural Information Processing Systems}, 34, 2021.

\bibitem[Jubran et~al.(2021)Jubran, Sanches~Shayda, Newman, and
  Feldman]{jubran2021coresets}
Jubran, I., Sanches~Shayda, E.~E., Newman, I., and Feldman, D.
\newblock Coresets for decision trees of signals.
\newblock \emph{Advances in Neural Information Processing Systems}, 34, 2021.

\bibitem[Karamcheti et~al.(2021)Karamcheti, Krishna, Fei-Fei, and
  Manning]{karamcheti2021mind}
Karamcheti, S., Krishna, R., Fei-Fei, L., and Manning, C.~D.
\newblock Mind your outliers! investigating the negative impact of outliers on
  active learning for visual question answering.
\newblock \emph{arXiv preprint arXiv:2107.02331}, 2021.

\bibitem[Killamsetty et~al.(2021)Killamsetty, Sivasubramanian, Mirzasoleiman,
  Ramakrishnan, De, and Iyer]{killamsetty2021grad}
Killamsetty, K., Sivasubramanian, D., Mirzasoleiman, B., Ramakrishnan, G., De,
  A., and Iyer, R.
\newblock Grad-match: A gradient matching based data subset selection for
  efficient learning.
\newblock \emph{arXiv preprint arXiv:2103.00123}, 2021.

\bibitem[Kothawade et~al.(2021)Kothawade, Beck, Killamsetty, and
  Iyer]{kothawade2021similar}
Kothawade, S., Beck, N., Killamsetty, K., and Iyer, R.
\newblock Similar: Submodular information measures based active learning in
  realistic scenarios.
\newblock \emph{Advances in Neural Information Processing Systems}, 34, 2021.

\bibitem[Kothyari et~al.(2021)Kothyari, Mekala, Iyer, Ramakrishnan, and
  Jyothi]{kothyari2021personalizing}
Kothyari, M., Mekala, A.~R., Iyer, R., Ramakrishnan, G., and Jyothi, P.
\newblock Personalizing asr with limited data using targeted subset selection.
\newblock \emph{arXiv preprint arXiv:2110.04908}, 2021.

\bibitem[Lai et~al.(2021)Lai, Zhang, Liu, Chang, Liao, Chuang, Qian, Khurana,
  Cox, and Glass]{lai2021parp}
Lai, C.-I.~J., Zhang, Y., Liu, A.~H., Chang, S., Liao, Y.-L., Chuang, Y.-S.,
  Qian, K., Khurana, S., Cox, D., and Glass, J.
\newblock Parp: Prune, adjust and re-prune for self-supervised speech
  recognition.
\newblock \emph{arXiv preprint arXiv:2106.05933}, 2021.

\bibitem[Liu et~al.(2022)Liu, Hsu, Auli, and Baevski]{liu2022towards}
Liu, A.~H., Hsu, W.-N., Auli, M., and Baevski, A.
\newblock Towards end-to-end unsupervised speech recognition.
\newblock \emph{arXiv preprint arXiv:2204.02492}, 2022.

\bibitem[Margatina et~al.(2021)Margatina, Vernikos, Barrault, and
  Aletras]{margatina-etal-2021-active}
Margatina, K., Vernikos, G., Barrault, L., and Aletras, N.
\newblock Active learning by acquiring contrastive examples.
\newblock In \emph{Proceedings of the 2021 Conference on Empirical Methods in
  Natural Language Processing}, pp.\  650--663, Online and Punta Cana,
  Dominican Republic, November 2021. Association for Computational Linguistics.

\bibitem[Mendon{\c{c}}a et~al.(2014)Mendon{\c{c}}a, Candeias, Perdigao, Shulby,
  Toniazzo, Klautau, and Alu{\'\i}sio]{mendoncca2014method}
Mendon{\c{c}}a, G., Candeias, S., Perdigao, F., Shulby, C., Toniazzo, R.,
  Klautau, A., and Alu{\'\i}sio, S.
\newblock A method for the extraction of phonetically-rich triphone sentences.
\newblock In \emph{2014 International Telecommunications Symposium (ITS)}, pp.\
   1--5. IEEE, 2014.

\bibitem[Mindermann et~al.(2022)Mindermann, Brauner, Razzak, Sharma, Kirsch,
  Xu, H{\"o}ltgen, Gomez, Morisot, Farquhar, et~al.]{mindermann2022prioritized}
Mindermann, S., Brauner, J.~M., Razzak, M.~T., Sharma, M., Kirsch, A., Xu, W.,
  H{\"o}ltgen, B., Gomez, A.~N., Morisot, A., Farquhar, S., et~al.
\newblock Prioritized training on points that are learnable, worth learning,
  and not yet learnt.
\newblock In \emph{International Conference on Machine Learning}, pp.\
  15630--15649. PMLR, 2022.

\bibitem[Mirzasoleiman et~al.(2020)Mirzasoleiman, Bilmes, and
  Leskovec]{mirzasoleiman2020coresets}
Mirzasoleiman, B., Bilmes, J., and Leskovec, J.
\newblock Coresets for data-efficient training of machine learning models.
\newblock In \emph{International Conference on Machine Learning}, pp.\
  6950--6960. PMLR, 2020.

\bibitem[Ni et~al.(2015{\natexlab{a}})Ni, Leung, Wang, Chen, and
  Ma]{ni2015unsupervised}
Ni, C., Leung, C.-C., Wang, L., Chen, N.~F., and Ma, B.
\newblock Unsupervised data selection and word-morph mixed language model for
  tamil low-resource keyword search.
\newblock In \emph{2015 IEEE International Conference on Acoustics, Speech and
  Signal Processing (ICASSP)}, pp.\  4714--4718. IEEE, 2015{\natexlab{a}}.

\bibitem[Ni et~al.(2015{\natexlab{b}})Ni, Wang, Liu, Leung, Lu, and
  Ma]{ni2015submodular}
Ni, C., Wang, L., Liu, H., Leung, C.-C., Lu, L., and Ma, B.
\newblock Submodular data selection with acoustic and phonetic features for
  automatic speech recognition.
\newblock In \emph{2015 IEEE International Conference on Acoustics, Speech and
  Signal Processing (ICASSP)}, pp.\  4629--4633. IEEE, 2015{\natexlab{b}}.

\bibitem[Ni et~al.(2016)Ni, Leung, Wang, Liu, Rao, Lu, Chen, Ma, and
  Li]{ni2016cross}
Ni, C., Leung, C.-C., Wang, L., Liu, H., Rao, F., Lu, L., Chen, N.~F., Ma, B.,
  and Li, H.
\newblock Cross-lingual deep neural network based submodular unbiased data
  selection for low-resource keyword search.
\newblock In \emph{2016 IEEE International Conference on Acoustics, Speech and
  Signal Processing (ICASSP)}, pp.\  6015--6019. IEEE, 2016.

\bibitem[Panayotov et~al.(2015)Panayotov, Chen, Povey, and
  Khudanpur]{panayotov2015librispeech}
Panayotov, V., Chen, G., Povey, D., and Khudanpur, S.
\newblock Librispeech: an asr corpus based on public domain audio books.
\newblock In \emph{2015 IEEE international conference on acoustics, speech and
  signal processing (ICASSP)}, pp.\  5206--5210. IEEE, 2015.

\bibitem[Paul et~al.(2021)Paul, Ganguli, and Dziugaite]{paul2021deep}
Paul, M., Ganguli, S., and Dziugaite, G.~K.
\newblock Deep learning on a data diet: Finding important examples early in
  training.
\newblock \emph{Advances in Neural Information Processing Systems}, 34, 2021.

\bibitem[Raju et~al.(2021)Raju, Daruwalla, and Lipasti]{raju2021accelerating}
Raju, R.~S., Daruwalla, K., and Lipasti, M.
\newblock Accelerating deep learning with dynamic data pruning.
\newblock \emph{arXiv preprint arXiv:2111.12621}, 2021.

\bibitem[Schirrmeister et~al.(2022)Schirrmeister, Liu, Hooker, and
  Ball]{schirrmeister2022less}
Schirrmeister, R.~T., Liu, R., Hooker, S., and Ball, T.
\newblock When less is more: Simplifying inputs aids neural network
  understanding.
\newblock \emph{arXiv preprint arXiv:2201.05610}, 2022.

\bibitem[Shim et~al.(2021)Shim, Choi, and Sung]{shim2021understanding}
Shim, K., Choi, J., and Sung, W.
\newblock Understanding the role of self attention for efficient speech
  recognition.
\newblock In \emph{International Conference on Learning Representations}, 2021.

\bibitem[Sorscher et~al.(2022)Sorscher, Geirhos, Shekhar, Ganguli, and
  Morcos]{sorscher2022beyond}
Sorscher, B., Geirhos, R., Shekhar, S., Ganguli, S., and Morcos, A.~S.
\newblock Beyond neural scaling laws: beating power law scaling via data
  pruning.
\newblock \emph{arXiv preprint arXiv:2206.14486}, 2022.

\bibitem[Thomas et~al.(2022)Thomas, Kessler, and Karout]{thomas2022efficient}
Thomas, B., Kessler, S., and Karout, S.
\newblock Efficient adapter transfer of self-supervised speech models for
  automatic speech recognition.
\newblock In \emph{ICASSP 2022-2022 IEEE International Conference on Acoustics,
  Speech and Signal Processing (ICASSP)}, pp.\  7102--7106. IEEE, 2022.

\bibitem[Tolochinsky \& Feldman(2018)Tolochinsky and
  Feldman]{tolochinsky2018coresets}
Tolochinsky, E. and Feldman, D.
\newblock Coresets for monotonic functions with applications to deep learning.
\newblock \emph{CoRR, abs/1802.07382}, 2018.

\bibitem[Toneva et~al.(2018)Toneva, Sordoni, Combes, Trischler, Bengio, and
  Gordon]{toneva2018empirical}
Toneva, M., Sordoni, A., Combes, R. T.~d., Trischler, A., Bengio, Y., and
  Gordon, G.~J.
\newblock An empirical study of example forgetting during deep neural network
  learning.
\newblock \emph{arXiv preprint arXiv:1812.05159}, 2018.

\bibitem[Wei et~al.(2014)Wei, Liu, Kirchhoff, Bartels, and
  Bilmes]{wei2014submodular}
Wei, K., Liu, Y., Kirchhoff, K., Bartels, C., and Bilmes, J.
\newblock Submodular subset selection for large-scale speech training data.
\newblock In \emph{2014 IEEE International Conference on Acoustics, Speech and
  Signal Processing (ICASSP)}, pp.\  3311--3315. IEEE, 2014.

\bibitem[Wolf et~al.(2019)Wolf, Debut, Sanh, Chaumond, Delangue, Moi, Cistac,
  Rault, Louf, Funtowicz, et~al.]{wolf2019huggingface}
Wolf, T., Debut, L., Sanh, V., Chaumond, J., Delangue, C., Moi, A., Cistac, P.,
  Rault, T., Louf, R., Funtowicz, M., et~al.
\newblock Huggingface's transformers: State-of-the-art natural language
  processing.
\newblock \emph{arXiv preprint arXiv:1910.03771}, 2019.

\bibitem[Woodard \& Nelson(1982)Woodard and Nelson]{woodard1982information}
Woodard, J. and Nelson, J.
\newblock An information theoretic measure of speech recognition performance.
\newblock In \emph{Workshop on standardisation for speech I/O technology, Naval
  Air Development Center, Warminster, PA}, 1982.

\bibitem[Wu et~al.(2007)Wu, Zhang, and Rudnicky]{wu2007data}
Wu, Y., Zhang, R., and Rudnicky, A.
\newblock Data selection for speech recognition.
\newblock In \emph{2007 IEEE Workshop on Automatic Speech Recognition \&
  Understanding (ASRU)}, pp.\  562--565. IEEE, 2007.

\bibitem[Wu et~al.(2021)Wu, Zhao, Liang, Yu, Gulati, and Pang]{wu2021dynamic}
Wu, Z., Zhao, D., Liang, Q., Yu, J., Gulati, A., and Pang, R.
\newblock Dynamic sparsity neural networks for automatic speech recognition.
\newblock In \emph{ICASSP 2021-2021 IEEE International Conference on Acoustics,
  Speech and Signal Processing (ICASSP)}, pp.\  6014--6018. IEEE, 2021.

\bibitem[Zhen et~al.(2021)Zhen, Nguyen, Chang, Mouchtaris, and
  Rastrow]{zhen2021sparsification}
Zhen, K., Nguyen, H.~D., Chang, F.-J., Mouchtaris, A., and Rastrow, A.
\newblock Sparsification via compressed sensing for automatic speech
  recognition.
\newblock In \emph{ICASSP 2021-2021 IEEE International Conference on Acoustics,
  Speech and Signal Processing (ICASSP)}, pp.\  6009--6013. IEEE, 2021.

\end{thebibliography}

%%%%%%%%%%%%%%%%%%%%%%%%%%%%%%%%%%%%%%%%%%%%%%%%%%%%%%%%%%%%%%%%%%%%%%%%%%%%%%%
%%%%%%%%%%%%%%%%%%%%%%%%%%%%%%%%%%%%%%%%%%%%%%%%%%%%%%%%%%%%%%%%%%%%%%%%%%%%%%%
% APPENDIX
%%%%%%%%%%%%%%%%%%%%%%%%%%%%%%%%%%%%%%%%%%%%%%%%%%%%%%%%%%%%%%%%%%%%%%%%%%%%%%%
%%%%%%%%%%%%%%%%%%%%%%%%%%%%%%%%%%%%%%%%%%%%%%%%%%%%%%%%%%%%%%%%%%%%%%%%%%%%%%%
\newpage
\appendix
\onecolumn

\section{Implementation Details}
\label{section:implementationdetails}
\subsection{Resources} 
\label{section:resources}
We use a single 80GB NVIDIA A100 GPU for running all the experiments on the cloud. In this setting, the standard \texttt{wav2vec2-base} fine-tuning step (single run) on multiple pruning fractions took $\approx1.25$ GPU hours for the TIMIT dataset, $\approx6$ GPU hours for LJSpeech dataset, and $\approx5.5$ GPU hours for Librispeech 10h dataset. The total project (from the early experiments to the final results) consumed about 2200 GPU hours.

\subsection{Code and Licenses}
\label{section:codeappendix}
We release our code under the MIT license. All the data pruning strategies are implemented in Python, and the resulting subsets are used to fine-tune \texttt{wav2vec2}. The publicly available HuggingFace \citep{wolf2019huggingface} implementation \footnote{https://github.com/huggingface/transformers} of \texttt{wav2vec2-base}  model\footnote{https://huggingface.co/facebook/wav2vec2-base-960h} is used which is based on the standard \texttt{wav2vec2-base-960h} fairseq implementation\footnote{https://github.com/pytorch/fairseq/blob/main/examples/wav2vec/README.md}. The HuggingFace transformers repo is available under the Apache License 2.0 license and the fairseq repo is available under the MIT license.

\subsection{Data}
\label{section:dataappendix}
\textbf{TIMIT} \citep{garofolo1993darpa}. We use the full TIMIT dataset  with predefined training and test sets. The training set contains 4620 examples and the test set contains 1680 examples. TIMIT is available under the LDC User Agreement for Non-Members. \\
\textbf{Librispeech} \citep{panayotov2015librispeech}. We construct Librispeech 10h fine-tuning split by selecting 10h of utterances randomly from the 100h train-clean split. The test-clean split is used for evaluation. Librispeech is available under the CC BY 4.0 license. \\
\textbf{LJSpeech} \citep{ito2017lj}. This dataset contains 24 hours of English speech from a single speaker. For validation and testing, we randomly select 300 utterances, mirroring the protocol followed in earlier works \cite{liu2022towards}. The rest is used for training. LJSpeech is available under the public domain license.

\subsection{Training}
\label{section:trainingdetails}
In all experiments, \texttt{wav2vec2-base} is fine-tuned with a batch size = 8, epochs = 20, mean ctc-loss-reduction, weight decay 0.005, and FP16 training. We use a data collator to pad the inputs dynamically. For calculating the WER for each training example, we run a computation step after each epoch and record the WER. The training WER in each epoch is averaged over 10 runs and then used for a particular pruning strategy. For each test WER reported, we do three separate runs with independent model initialization. A bucket size of $500$ is chosen for the \textsc{Cowerage} strategy, which is sufficiently small to ensure the selection of representative examples for different pruning fractions.

\section{Additional Experiments}
\label{section:additionalexperiments}

\subsection{WER Selection Epoch}
\label{section:werselectionepoch}
An important hyperparameter in the \textsc{Cowerage} algorithm is the epoch at which the training WER is computed for individual examples and then used for pruning i.e. the WER selection epoch. We evaluate the effect of different selection epochs on the final test WER (Table \ref{tab:werselectionepoch}) in TIMIT and observe that the training WER in the early training epochs can be reliably used for ranking the examples and applying a particular pruning strategy. Hence, we select WSE = 8 for the final results in Table \ref{tab:pruningresults}. Note that \textsc{Cowerage} consistently demonstrates a lower WER than other strategies on \textit{all epochs} that we test (8, 12, 16, 20) for the majority of pruning fractions ($0.2-0.9$) across all the datasets (TIMIT, LS-10h, LJSpeech). This suggests that the selection of a reasonable WSE can usually be made with less than five distinct epoch values while still achieving better results than the other strategies.

\begin{table*}[!htbp]

\centering
\caption{Test WER for the four strategies of pruning the training set evaluated at multiple pruning fractions and different training WER selection epochs. The training WER in a particular selection epoch is averaged over 10 runs and then used for a particular pruning strategy. For each result, we do three independent runs and report the mean test WER. \textsc{Cowerage} consistently demonstrates the lowest WER at various pruning fractions and selection epochs. WSE: WER Selection Epoch.}

\resizebox{0.7\textwidth}{!}{%
\begin{tabular}{c c c c c c c c c c c}
%\small
\toprule
WSE &  Strategy & \multicolumn{6}{c}{Pruning Fraction}\\
\midrule
{} & {} & \textbf{No pruning} & \textbf{0.1} & \textbf{0.3} & \textbf{0.5} & \textbf{0.7} & \textbf{0.9} \\
\midrule
8    &  Random            &  0.315 &  0.325 & 0.341 & 0.357 & 0.394 & 0.557\\
{}   &  Top K             & 0.315 & 0.322 &		0.334 &		0.392 &	0.472 &	0.678 \\
{}   &  Bottom K          & 0.315 & 0.336  &	0.360  &	0.411 &	0.521 & 	0.887 \\
{}   &  \textsc{Cowerage} & 0.315 & \textbf{0.320}   &	 \textbf{0.333}  &	\textbf{0.339}		& \textbf{0.369} &	\textbf{0.455}  \\
\midrule
12   &  Random            & 0.315 &  0.325 & 0.341 & 0.357 & 0.394 & 0.557 \\
{}   &  Top K             & 0.315 &  0.316 & 0.345 & 0.386 & 0.461 & 0.579 \\
{}   &  Bottom K          & 0.315 & 0.323 & 0.353 & 0.398 & 0.499 & 0.781 \\
{}   &  \textsc{Cowerage} & 0.315 & \textbf{0.322} & \textbf{0.328} & \textbf{0.354} & \textbf{0.370} & \textbf{0.536}  \\
\midrule
16   &  Random            & 0.315 &  0.325 & 0.341 & 0.357 & 0.394 & 0.557 \\
{}   &  Top K             & 0.315 &  0.324 & 0.332 & 0.413 & 0.467 & 0.704 \\
{}   &  Bottom K          & 0.315 &  0.323 & 0.346 & 0.382 & 0.468 & 0.657 \\
{}   &  \textsc{Cowerage} & 0.315 &  \textbf{0.322} & \textbf{0.329} & \textbf{0.356} & \textbf{0.382} & \textbf{0.565}   \\
\midrule
20   &  Random            & 0.315 &  0.324 & 0.340 & 0.357 & 0.401 & 0.557 \\
{}   &  Top K             & 0.315 &  0.328 & 0.370 & 0.422 & 0.518 & 0.709 \\
{}   &  Bottom K          & 0.315 &  0.321 & 0.352 & 0.389 & 0.457 & 0.587 \\
{}   &  \textsc{Cowerage} & 0.315 &  \textbf{0.321} & \textbf{0.334} & \textbf{0.340} & \textbf{0.376} & \textbf{0.545}   \\

\bottomrule
\end{tabular}}
\vskip 0.2in
\label{tab:werselectionepoch}
\end{table*}

\subsection{Training WER Distribution}
\label{section:trainingwerdistribution}
\begin{figure*}[!ht]
\centering
  \includegraphics[width=0.9\textwidth]{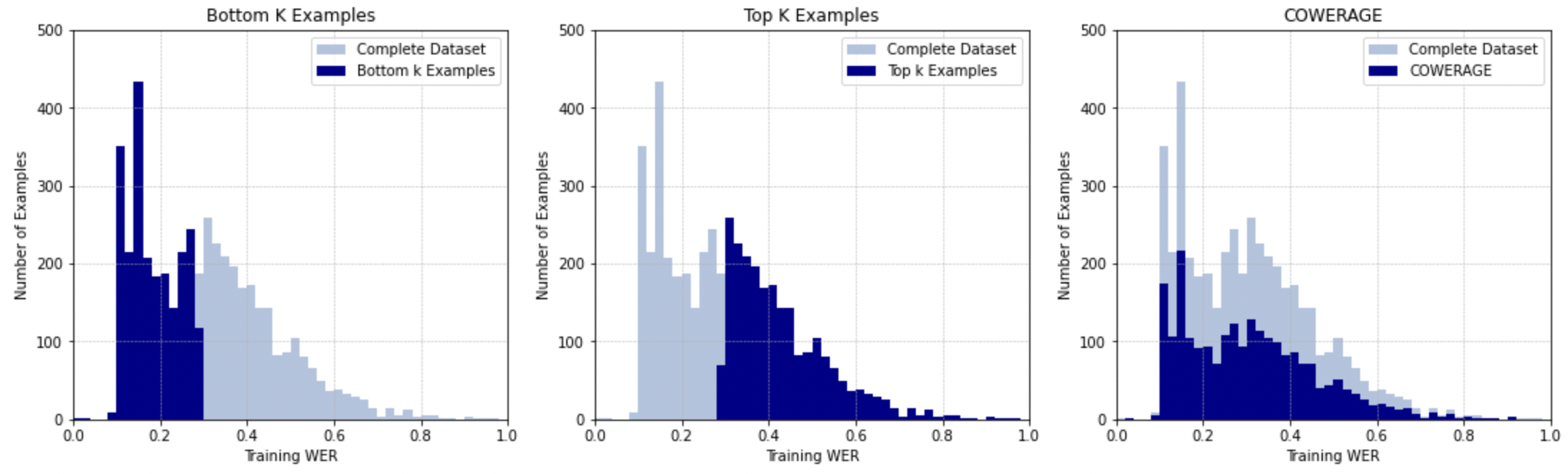}
  \caption{The subset of the TIMIT training data selected by each of the three strategies: bottom \textit{k} (\textit{left}), top \textit{k} (\textit{middle}) and \textsc{Cowerage} (\textit{right}). The pruning fraction is set to 0.5 and the WER selection epoch is 8.}
  \label{fig:timitdistribution}
  \vskip -0.1in
\end{figure*}

\begin{figure*}[!ht]
\centering
  \includegraphics[width=0.9\textwidth]{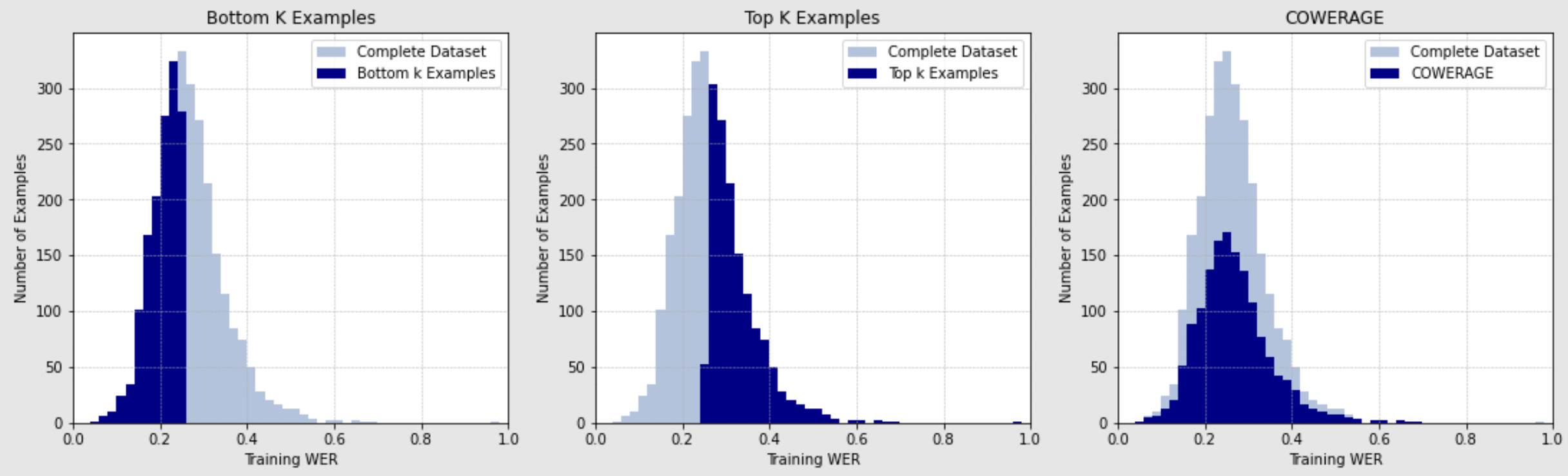}
  \caption{The subset of the Librispeech 10h training data selected by each of the three strategies: bottom \textit{k} (\textit{left}), top \textit{k} (\textit{middle}) and \textsc{Cowerage} (\textit{right}). The pruning fraction is set to 0.5 and the WER selection epoch is 8.}
  \label{fig:lsdistribution}
  \vskip -0.1in
\end{figure*}

\begin{figure*}[!ht]
\centering
  \includegraphics[width=0.9\textwidth]{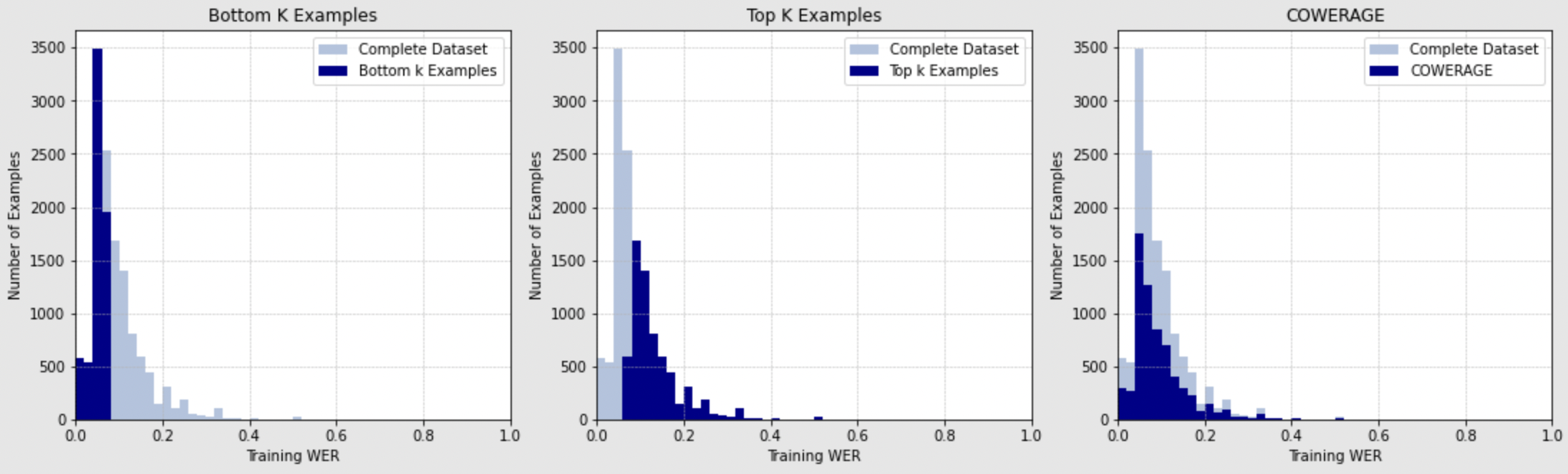}
  \caption{The subset of the LJspeech 24h training data selected by each of the three strategies: bottom \textit{k} (\textit{left}), top \textit{k} (\textit{middle}) and \textsc{Cowerage} (\textit{right}). The pruning fraction is set to 0.5 and the WER selection epoch is 8.}
  \label{fig:ljdistribution}
  \vskip -0.1in
\end{figure*}

We compare the distribution of the training WER for TIMIT  (Fig. \ref{fig:timitdistribution}), Librispeech 10h (Fig. \ref{fig:lsdistribution}) and LJSpeech (Fig. \ref{fig:ljdistribution}) and show the subsets selected through Top K, Bottom K and \textsc{Cowerage} subset selection on 50\% pruning percentage.  We notice significant differences in the training WER distribution for the three datasets which highlights that the example difficulty (measured by WER) is a property of the dataset. Moreover, since \textsc{Cowerage} performs better than other subset selection methods across multiple datasets, we hypothesize that the our proposed method is dataset-agnostic and can perform well with different training WER distributions.

\subsection{Training Landscape}

\begin{figure*}[!ht]
\centering
  \includegraphics[width=1\textwidth]{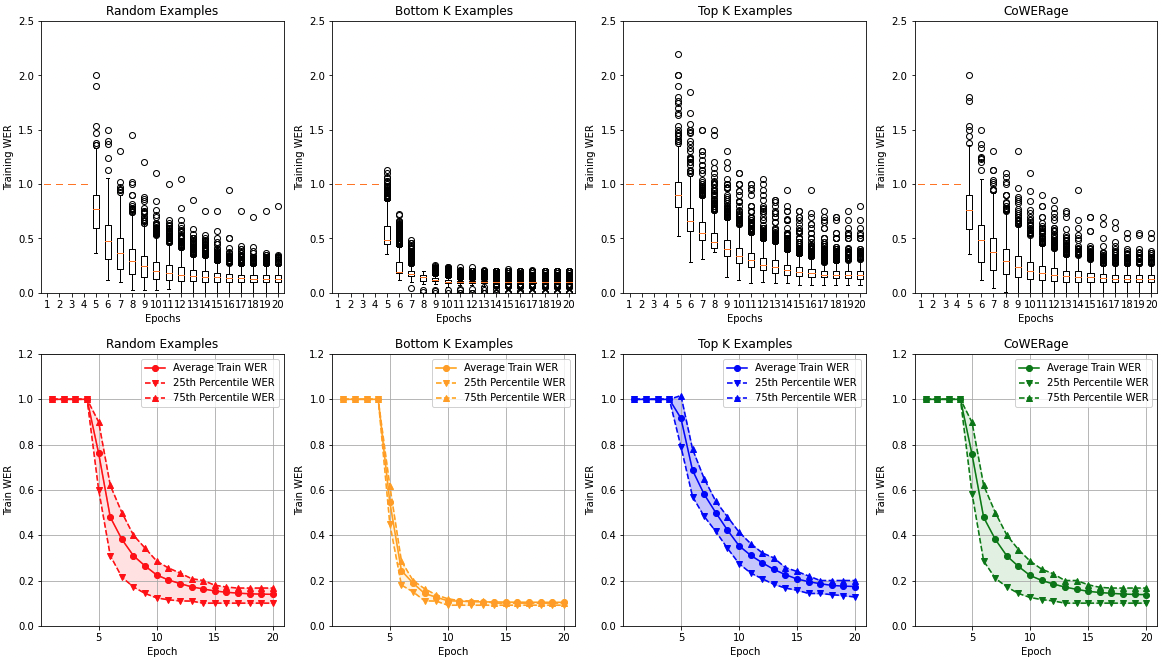}
  \caption{The training trajectories of the examples in TIMIT selected by picking the random examples (\textit{first column}),  bottom k examples (\textit{second column}), top k examples (\textit{third column}), and via \textsc{Cowerage} subset selection (\textit{fourth column}). For each epoch, we show the box plot of the distribution of the word error rate of training examples that indicates the mean, quartiles, and outliers.  }
  \label{fig:traintrajectories}
\end{figure*}

We now compare the training landscape for the three strategies discussed. We create four subsets of data at the pruning fraction of 0.7 and plot the training WER for each of the four approaches (Fig. \ref{fig:traintrajectories}). By examining the outlier behavior and the width of the box plots (25th to 75th percentile), we find that \textsc{Cowerage} subset selection is actually picking the \textit{moderately hard} and \textit{representative} examples instead of just the \textit{very hard} but \textit{rare} examples. 

\subsection{Selecting the number of buckets}
\label{section:selectingbuckets}
We conduct an experiment with different bucket sizes on \texttt{wav2vec2} and TIMIT with 0.7 pruning fraction. The results are shown in Table \ref{tab:bucketsexperiment}. Our evaluation shows that increasing the bucket size beyond a certain threshold provides diminishing returns in performance. Increasing the bucket size from 50 to 100 yielded 4.8\% reduction in WER whereas increasing it from 100 to 500 resulted in only a 0.27\% reduction in WER.
    
    \begin{table}[!hptb]
    \centering
    \caption{Test WER for \texttt{wav2vec2} on TIMIT for different number of buckets in the \textsc{Cowerage} algorithm}
    
    \begin{tabular}{ c c c c c c c}
    \toprule
    \textbf{Number of Buckets} & 1 & 10 & 50 & 100 & 500 & 1000 \\
    \midrule
    \textbf{Test WER} & 0.394 & 0.393 &	0.389 &	0.370 &	0.369 & 0.369 \\
    \bottomrule
    \end{tabular}
    \label{tab:bucketsexperiment}
    \end{table}
    
Choosing 500 buckets in the \textsc{Cowerage} algorithm provided robust performance across a wide range of dataset sizes, which ranged from 4620 examples in TIMIT to more than 10,000 examples in LJSpeech.
The number of buckets can be increased further but it should be no greater than \texttt{pruningFraction * datasetSize}.

\subsection{Transferability to larger models}
\label{section:largermodels}

To find out if the subsets created through a smaller model are transferable to a larger speech SSL model, we conduct an experiment with \texttt{wav2vec2-large} (317M parameters; pre-trained on Librispeech 960h) and fine-tune it on the subsets constructed through \texttt{wav2vec2-base}. We observe that \textsc{Cowerage} subsets still outperform the rest of the pruning strategies, further validating the hypothesis of transferability of pruning scores.

\begin{table}[!htbp]
\centering
\caption{Test WER for different for \texttt{wav2vec2-large} fine-tuned on subsets created through \texttt{wav2vec2-base}.}
\vskip 0.1in

\resizebox{0.5\textwidth}{!}{%
\begin{tabular}{c c c c c c c}
%\small
\toprule
Strategy & \multicolumn{5}{c}{Pruning Fraction}\\
\midrule
\midrule
{} & \textbf{0.1} & \textbf{0.3} & \textbf{0.5}  & \textbf{0.7}  & \textbf{0.9} \\
\midrule
Random  &  0.300 & 	0.308 &		0.322 &	0.356	  & 	0.545 \\
Top K  & 0.295  &  0.297 & 0.345 & 	0.385 &	0.634	\\
Bottom K & 0.306  &	0.326  &	0.391  &	0.505 &	0.833 \\
\textsc{Cowerage} & \textbf{0.290}  &	\textbf{0.296}  &	\textbf{0.318}		& \textbf{0.332}   & \textbf{0.490}	\\

\bottomrule
\end{tabular}}
\label{tab:phonemerecognition}
\vskip 1in
\end{table}

\subsection{Standard deviation for test WER on TIMIT}

\begin{table*}[htpb]
\centering
\caption{The standard deviation for the test WER of \texttt{wav2vec2} presented in Table. \ref{tab:werselectionepoch}}

\resizebox{0.7\textwidth}{!}{%
\begin{tabular}{c c  c c c c c}
\toprule
WSE &  Strategy & \multicolumn{5}{c}{Pruning Fraction}\\
\midrule
\midrule
{} & {} & \textbf{0.1}  & \textbf{0.3}  & \textbf{0.5}  & \textbf{0.7} & \textbf{0.9} \\

\midrule
TIMIT    &  Random  & $ \pm 0.003 $  &	$ \pm 0.005 $  &	$ \pm 0.002 $ &	$ \pm 0.025 $ &	$ \pm 0.003 $ \\
{}   &  Top K  & $ \pm 0.001 $  &	$ \pm 0.007 $  &	$ \pm 0.010 $ &	$ \pm 0.001 $ &	$ \pm 0.002 $\\
{}   &  Bottom K & $ \pm 0.002 $ &	$ \pm 0.002 $ &	$ \pm 0.002 $  &	$ \pm 0.009 $  &	$ \pm 0.002 $ \\
{}   & \textsc{Cowerage} &  $ \pm 0.001 $  &	$ \pm 0.006 $ &	$ \pm 0.016 $  &	$ \pm 0.004 $ &		$ \pm 0.005 $\\

\bottomrule
\end{tabular}}
\label{tab:pruningresultsstd}
\end{table*}

\subsection{Length and Phonemes }
\label{section:lengthvsphonemes}
\begin{figure*}[h]
  \centering
  \includegraphics[width=0.4\textwidth]{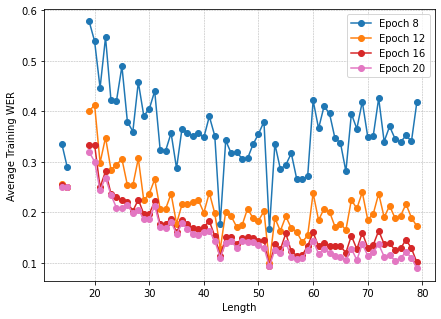}
  \caption{The training WER and the length of the examples (total number of characters) in TIMIT dataset compared over multiple training epochs. The WER is computed by averaging the training scores of the examples with the same length.}
  \label{fig:phonemevswerfig}
\end{figure*}

In this section, we examine the relationship between length and the training WER and conduct the same experiment from Section \ref{section:connectiontophonemes} but now with the length instead of the phonemic cover. The results are shown in Figure \ref{fig:phonemevswerfig}. The overall inverse relationship is similar to the one in Figure \ref{fig:trainingwervsphonemiccover} but is noisier. We notice that there are shorter and longer sentences with a high training WER in the earlier training epochs. If we bucket the examples by length, each bucket has a higher variance of WER values than the phoneme experiment in Figure \ref{fig:trainingwervsphonemiccover}. We also evaluate a variant of \textsc{Cowerage} that selects examples on the basis of their character length instead of WER which demonstrates that WER sampling is a better subset selection strategy than length sampling for the majority of pruning fractions (Table \ref{tab:coweragelength}).

\begin{table*}[!htbp]

\centering
\caption{Test WER for a variant of \textsc{Cowerage} that selects examples based on their length instead of WER. } 

\resizebox{0.6\textwidth}{!}{%
\begin{tabular}{c c c c c c c}
%\small
\toprule
Model &  Strategy & \multicolumn{5}{c}{Pruning Fraction}\\
\midrule
\midrule
{} & {} & \textbf{0.1} & \textbf{0.3} & \textbf{0.5} & \textbf{0.7} & \textbf{0.9} \\
\midrule
\texttt{wav2vec2-base} 
{}   &  \textsc{Cowerage} (Length) & \textbf{0.318} & \textbf{0.323} & 0.366 & 0.399 & 0.587 \\
{}   &  \textsc{Cowerage} (WER) & 0.320   &	 0.333  &	\textbf{0.339}		& \textbf{0.369} &	\textbf{0.455}

\end{tabular}}
\label{tab:coweragelength}
\end{table*}

\clearpage

\subsection{Examples}
\label{trainandtestexamples}

\begin{table*}[!htpb]
\centering
\caption{Training examples in the TIMIT dataset and their training WER on \texttt{wav2vec2} along with the phonemic cover (PC). The training WER is calculated by averaging 10 runs.}

\resizebox{0.9\textwidth}{!}{%
\begin{tabular}{p{1cm} p{4cm} p{9cm} p{1cm} }
\toprule
WER &  Text& Phonemes & PC \\
\midrule
\midrule

0.63   &  Twelve o'clock level. & (t-w-eh-l-v-ax-kcl-k-l-aa-kcl-k-l-eh-v-el) & 10 \\
0.63   &  That's your headache. & (dh-ae-tcl-t-s-y-er-hv-eh-dx-ey-kcl-k) & 13 \\
0.6   &  Run-down, iron-poor. & (r-ah-n-dcl-d-aw-n-q-ay-er-n-pcl-p-ao-r
) & 12\\
0.49   &  Y'all wanna walk -- walk, he said. & (y-ao-l-w-ao-n-ax-w-ao-kcl-pau-w-ao-kcl-k-iy-s-eh-dcl
) & 13\\
0.46   &  Pansies are gluttons. & (p-ae-n-z-iy-z-er-gcl-g-l-ah-tcl-en-d-z
) & 13\\
0.43   &  She seemed irritated. & (sh-iy-s-ey-m-dcl-d-ih-er-tcl-t-ey-dx-ix-dcl) & 13\\
0.42   &  Where're you takin' me? & (w-er-y-ux-tcl-t-ey-kcl-k-ix-n-m-iy) & 13\\
0.41   &  They're doin' it now. & (dh-eh-r-dcl-d-uw-ih-nx-ih-tcl-n-aw) & 11\\
0.40   &  Yes, ma'am, it sure was. & (y-eh-s-epi-m-ae-m-ih-tcl-t-sh-er-w-ah-s) & 13\\
0.40   &  Twenty-two or twenty-three. & (t-w-eh-n-tcl-t-iy-tcl-t-ux-ao-r-tcl-t-w-eh-n-tcl-t-iy-th-r-iy) & 10\\

\midrule
0.07   &  Boys and men go along the riverbank or to the alcoves in the top arcade. & (b-oy-z-ix-n-m-eh-n-gcl-g-ow-ax-l-ao-ng-n-ix-r-ih-v-er-bcl-b-ae-ng-kcl-k-q-ao-r-tcl-t-ux-dcl-d-iy-q-ae-l-kcl-k-ow-v-z-q-ix-n-dh-ix-tcl-t-aa-pcl-p-aa-r-kcl-k-ey-dcl-d) & 34 \\

0.07   &  But if she wasn't interested, she'd just go back to the same life she'd left. & (b-uh-dx-ih-f-sh-iy-w-ah-z-ix-n-ih-n-tcl-t-axr-s-tcl-t-ih-dcl-d-pau-sh-iy-dcl-jh-uh-s-gcl-g-ow-bcl-b-ae-kcl-t-ix-dh-ix-s-ey-m-l-ay-f-sh-iy-dcl-l-eh-f-tcl-t) & 32 \\

0.07   &  Why the hell didn't you come out when you saw them gang up on me? & (w-ay-dh-eh-hv-eh-l-dcl-d-ih-dcl-en-tcl-ch-ux-kcl-k-ah-m-aw-q-w-ix-n-y-ux-s-ao-dh-ix-m-gcl-g-ae-ng-ah-pcl-p-ao-n-m-iy) & 31 \\

0.06   &  You think somebody is going to stand up in the audience and make guilty faces? & (y-ux-th-ih-ng-kcl-k-s-ah-m-bcl-b-aa-dx-iy-ix-z-gcl-g-oy-ng-dcl-d-ix-s-tcl-t-ae-n-dcl-d-ah-pcl-p-ix-n-ah-q-aa-dx-iy-eh-n-tcl-s-eh-m-ey-kcl-g-ih-l-tcl-t-ix-f-ey-s-eh-z) & 33 \\

0.06   &  How much and how many profits could a majority take out of the losses of a few? & (hh-aw-m-ah-tcl-ch-ix-n-hv-aw-m-ax-nx-iy-pcl-p-r-aa-f-ax-tcl-s-kcl-k-uh-dx-ax-m-ax-dcl-jh-ao-axr-dx-iy-tcl-t-ey-kcl-k-ae-dx-ah-dh-ax-l-ao-s-ix-z-ax-v-ax-f-y-ux) & 35 \\

0.06   &  He may not rise to the heights, but he can get by, and eventually be retired. & (hh-iy-m-ey-n-aa-q-r-ay-z-tcl-t-ix-dh-ax-hv-ay-tcl-s-pau-b-ah-dx-iy-kcl-k-ix-ng-gcl-g-eh-q-bcl-b-ay-pau-q-ix-nx-iy-v-eh-n-ch-ix-l-iy-pau-b-iy-r-iy-tcl-t-ay-axr-dcl-d) & 35 \\

0.06   &  My sincere wish is that he continues to add to this record he sets here today. & (m-ay-s-en-s-ih-r-w-ih-sh-ix-z-dh-eh-tcl-hv-iy-kcl-k-ax-h-tcl-t-ih-n-y-ux-z-tcl-t-ax-h-q-ae-dcl-d-pau-t-ux-dh-ih-sh-r-eh-kcl-k-axr-dx-iy-s-eh-tcl-s-hh-ix-r-tcl-t-ax-h-dx-ey) & 31 \\

0.05   &  Then he fled, not waiting to see if she minded him or took notice of his cry. & (dh-ih-n-iy-f-l-eh-dcl-d-pau-n-aa-q-w-ey-dx-ih-ng-dcl-d-ix-s-iy-ih-f-sh-iy-m-ay-n-ix-dcl-d-hv-ih-m-pau-q-axr-tcl-t-uh-kcl-n-ow-dx-ih-s-ix-v-ix-z-kcl-k-r-ay) & 32 \\

0.01   &  We apply auditory modeling to computer speech recognition. & (w-iy-ax-pcl-p-l-ay-q-ao-dx-ix-tcl-t-ao-r-ix-m-aa-dx-el-ix-ng-tcl-t-uw-kcl-k-ax-m-pcl-p-y-ux-dx-er-s-pcl-p-iy-tcl-ch-epi-r-eh-kcl-k-ix-gcl-n-ih-sh-ix-n) & 35\\

\bottomrule
\end{tabular}}
\label{tab:examples}
\end{table*}

\end{document}